\documentclass[10pt,journal,compsoc]{IEEEtran}
\usepackage{amsmath,amsfonts}
\usepackage{algorithmic}
\usepackage{algorithm}
\usepackage{array}
\usepackage[caption=false,font=normalsize,labelfont=sf,textfont=sf]{subfig}
\usepackage{textcomp}
\usepackage{stfloats}
\usepackage{url}
\usepackage{verbatim}
\usepackage{graphicx}
\usepackage{cite}
\usepackage{tikz}
\usepackage{comment}
\usepackage{amsmath}
\usepackage{amsfonts}
\usepackage{bbding}
\usepackage{pgfplots}
\usepackage{booktabs}
\usepackage{diagbox}
\usepackage{makecell}
\usepackage{multirow}
\usepackage{xcolor}  % for \xcolor{} in words
\usepackage{hhline}
\usepackage{colortbl}  % for \rowcolor{} in table line
\newcommand{\etal}{\textit{et al.}}
\newcommand{\eg}{\textit{e.g.}}
\newcommand{\ie}{\textit{i.e.}}
\usepackage{amssymb}% http://ctan.org/pkg/amssymb
\usepackage{pifont}% http://ctan.org/pkg/pifont
\newcommand{\cmark}{\ding{51}}%
\newcommand{\xmark}{\ding{55}}%

\usepackage[hidelinks]{hyperref}
\hypersetup{
    colorlinks,
    linkcolor={red!100!black},
    citecolor={green!100!black},
    urlcolor={blue!100!black}
}

\begin{document}
\title{BPJDet: Extended Object Representation for Generic Body-Part Joint Detection}

\author{Huayi Zhou, \IEEEmembership{Student Member, IEEE}, Fei Jiang, Jiaxin Si, Yue Ding, and Hongtao Lu, \IEEEmembership{Member, IEEE}
        % <-this % stops a space
\thanks{H. Zhou, Y. Ding, and H. Lu are with Department of Computer Science and Engineering, Shanghai Jiao Tong University, Shanghai 200240, China (e-mail: sjtu\_zhy@sjtu.edu.cn; dingyue@sjtu.edu.cn; htlu@sjtu.edu.cn). F. Jiang (corresponding author) is with Intelligent of AI Education, East China Normal University, Shanghai 200062, China (e-mail: fjiang@mail.ecnu.edu.cn). J. Si is working in Chongqing Qulian Digital Technology Company, China (e-mail: sijiaxin@hyperchain.cn). This paper is supported by NSFC (No. 62176155, 62207014), Shanghai Municipal Science and Technology Major Project (2021SHZDZX0102). Hongtao Lu is also with MOE Key Lab of Artificial Intelligence, AI Institute, Shanghai Jiao Tong University.}
\thanks{Manuscript received April 19, 2023; revised November 09, 2023; Accepted January 13, 2024.}}

\markboth{Journal of \LaTeX\ Class Files,~Vol.~14, No.~8, January~2024}%
{Shell \MakeLowercase{\textit{et al.}}: Body-Part Joint Detection and Association }

%\IEEEpubid{\copyright~2023 IEEE \qquad\qquad\qquad}
\IEEEpubid{0000--0000/00\$00.00~\copyright~2024 IEEE}

\IEEEtitleabstractindextext{

%\maketitle
%%%%%%%%%%%%%%%%%%%%%%%%%%%%%%%%%%%%%%%%%%%%%%%%%%%%%%%%%%%%%%%
\begin{abstract}
Detection of human body and its parts has been intensively studied. However, most of CNNs-based detectors are trained independently, making it difficult to associate detected parts with body. In this paper, we focus on the joint detection of human body and its parts. Specifically, we propose a novel extended object representation integrating center-offsets of body parts, and construct an end-to-end generic Body-Part Joint Detector (BPJDet). In this way, body-part associations are neatly embedded in a unified representation containing both semantic and geometric contents. Therefore, we can optimize multi-loss to tackle multi-tasks synergistically. Moreover, this representation is suitable for anchor-based and anchor-free detectors. BPJDet does not suffer from error-prone post matching, and keeps a better trade-off between speed and accuracy. Furthermore, BPJDet can be generalized to detect body-part or body-parts of either human or quadruped animals. To verify the superiority of BPJDet, we conduct experiments on datasets of body-part (CityPersons, CrowdHuman and BodyHands) and body-parts (COCOHumanParts and Animals5C). While keeping high detection accuracy, BPJDet achieves state-of-the-art association performance on all datasets. Besides, we show benefits of advanced body-part association capability by improving performance of two representative downstream applications: accurate crowd head detection and hand contact estimation. Project is available in \url{https://hnuzhy.github.io/projects/BPJDet}.
\end{abstract}

%%%%%%%%%%%%%%%%%%%%%%%%%%%%%%%%%%%%%%%%%%%%%%%%%%%%%%%%%%%%%%%

}
\maketitle
%%%%%%%%%%%%%%%%%%%%%%%%%%%%%%%%%%%%%%%%%%%%%%%%%%%%%%%%%%%%%%%
\section{Introduction}\label{intro}

\IEEEPARstart{H}{uman} body detection \cite{dollar2011pedestrian, zhang2018occlusion, wang2018repulsion, chi2020pedhunter, chu2020detection} is a long-standing research hotspot in the computer vision field. Accurate and fast human detection in an arbitrary scene can support many down-stream vision tasks such as pedestrian person re-identification (ReID) \cite{li2017learning, ye2021deep}, person tracking \cite{zhou2020tracking, sundararaman2021tracking, luo2021multiple} and human pose estimation \cite{he2017mask, fang2022alphapose}. Also, the detection of body parts like face \cite{hu2017finding, deng2020retinaface}, head \cite{vu2015context, le2018key} and hands \cite{zhou2018raising, narasimhaswamy2022whose} is equally important and has been extensively studied. They may serve as precursors to specific tasks like face recognition \cite{deng2019arcface}, crowd counting \cite{idrees2015detecting, sam2020locate} and hand pose estimation \cite{ge2018real}. Although the independent detection of human body and related parts has been significantly improved, due to breakthroughs of deep CNN-based general object detectors (e.g., Faster R-CNN \cite{ren2015faster}, FPN \cite{lin2017feature}, RetinaNet \cite{lin2017focal} and YOLO \cite{redmon2016you}) and construction of large-scale high-quality datasets (e.g., COCO \cite{lin2014microsoft}, CityPersons \cite{zhang2017citypersons} and CrowdHuman \cite{shao2018crowdhuman}), the more challenging joint discovery of human body and its parts is less-addressed but meaningful. As proof, recently, we can find numerous human-related applications excavating the linkage effect of body-part, including robot teleoperation \cite{gao2023parallel}, person surveillance \cite{xiong2019vikingdet, shu2021head}, robust person ReID \cite{ding2020multi, somers2023body}, human parsing \cite{zhang2022aiparsing, yang2022quality, liu2022cdgnet}, hand-related object contact \cite{brahmbhatt2020contactpose, narasimhaswamy2020detecting, shan2020understanding, muller2021self, chen2023detecting}, human-object interaction (HOI) \cite{li2020pastanet, wu2022mining, li2022hake, lim2023ernet}, part-based 3D human body models \cite{osman2022supr, su2022danbo, mihajlovic2022coap} and human pose estimation \cite{dantone2014body, zhang2020learning, kreiss2021openpifpaf, cao2021openpose}.

% face, head, right-hand/left-hand and left-foot/right-foot

\begin{figure}[!t]
	\centering
	\includegraphics[width=\columnwidth]{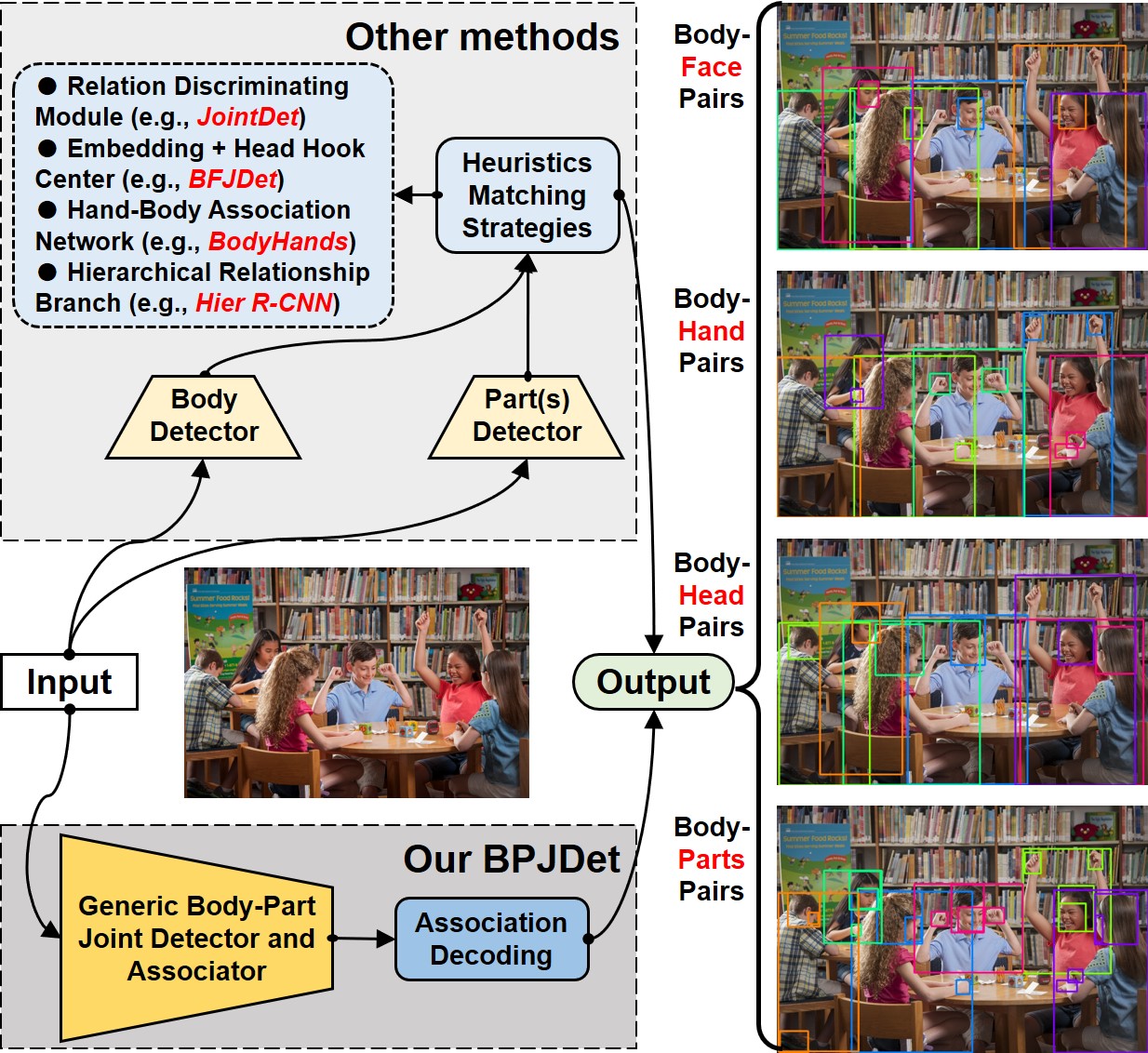}\\
	\vspace{-8pt}
	\caption{The illustration of the difference between our proposed single-stage BPJDet and other two-stage body-part joint detection methods (e.g., JointDet \cite{chi2020relational}, BFJDet \cite{wan2021body}, BodyHands \cite{narasimhaswamy2022whose} and Hier R-CNN \cite{yang2020hier}). Their two-stage refers to training the detection and association modules separately, unlike our one-stage joint detection and association framework. We visualize bodies and parts that belong to the same person using bounding boxes with the same color.}
	\label{illustrations}
	\vspace{-10pt}
\end{figure}

In this paper, we study the body-part joint detection task. A body part can be a face, head, hand or their arbitrary combinations as in Fig. \ref{illustrations}. Our principal goal is to improve the accuracy of body-part association on the premise of ensuring the detection precision. Prior to ours, a few literatures have attempted to address the joint body and part detection task with two separate stages. DA-RCNN \cite{zhang2019double}, PedHunter \cite{chi2020pedhunter} and JointDet \cite{chi2020relational} focus on joint body-head detection by assuming that human head can provide complementary spatial information to the occluded body. BFJDet \cite{wan2021body} concentrates on the similar body-face joint detection task which is much more difficult because body and face are not always one-to-one correspondence. BodyHands \cite{narasimhaswamy2022whose} and Zhou \etal \cite{zhou2018raising} tackle the problem of detecting hands with finding the location of corresponding person. Instead of only one body part, DID-Net \cite{li2019detector} takes the inherent correlation between the body and its two parts (hand and face) into account by building the dataset HumanParts. Hier R-CNN \cite{yang2020hier} extends HumanParts to COCOHumanParts based on COCO \cite{lin2014microsoft} and defines up to six body parts. However, as in Fig. \ref{illustrations}, two-stage methods use explicit strategies to model body-part relationships by exploiting heuristic post-matching rules or learning branched association networks. Unlike them, we propose a novel single-stage Body-Part Joint Detector (BPJDet), which can jointly detect and associate body with one or more parts in a unified framework. 

The association ability of proposed BPJDet mainly relies on our elaborately designed extended object representation (refer Fig. \ref{gridcell}) that can be harmlessly applied to single-stage anchor-based or anchor-free detectors like YOLO series \cite{redmon2016you, redmon2017yolo9000, wang2021scaled, yolov5, wang2022yolov7, xu2022pp, yolov8}. Besides bounding box, confidence and objectness contained in the classic object representation, our extension adds location offsets of body parts to it. Center-offset regression is popular in modern one-stage object detectors \cite{zhou2019objects, tian2019fcos} and human pose estimators \cite{cao2017realtime, nie2019single, mao2021fcpose, xiao2022adaptivepose}. This simple yet efficient innovation has at least three advantages. (1) The relational learning of body-part can benefit from training of general detectors without needing extra association subnets. (2) The final prediction naturally implies the detected body bounding boxes and their associated parts, which avoids error-prone and tedious association post-processing. (3) This general representation enables us to jointly detect arbitrary body part or parts such as face and hands without major modifications. In experiments, we performed extensive tests on many public datasets including CityPersons \cite{zhang2017citypersons}, CrowdHuman \cite{shao2018crowdhuman}, BodyHands \cite{narasimhaswamy2022whose} and COCOHumanParts \cite{yang2020hier} for human and the dataset Animals5C (built from AnimalPose \cite{cao2019cross} and AP-10K \cite{yu2021ap}) for quadruped animals to verify the superiority of our method. To further reflect the potential value of our advanced BPJDet, we have applied it to two closely related tasks: accurate crowd head detection and hand contact estimation.

To sum up, we mainly have following five contributions:
\begin{itemize}
\item We propose a novel end-to-end trainable Body-Part Joint Detector (BPJDet) by extending the classic object representation with offsets of body parts.
\item We reveal the feasibility and expedience of joint training of bounding boxes and offsets with designing suitable multi-task loss functions by using either anchor-based or anchor-free detectors.
\item We verify that BPJDet is a unified and universal framework which can support joint detection of body with one or more body parts for both human and some quadruped animals.
\item We achieve state-of-the-art body-part association performance on four public benchmarks while maintaining high detection accuracy of body and parts.
\item We demonstrate the significant benefits of BPJDet's superior body-part association ability for performance promotion in two downstream tasks.
\end{itemize}

In order to inspire the future research and application of generic body-part joint detection and association, we open source the implementation of BPJDet. In this work, we make several major extensions to our earlier conference paper \cite{zhou2022body}: (1) A more in-depth introduction to works related to the body-part joint detection task and its downstream applications in Sections \ref{intro} and \ref{relwork}. (2) A more detailed discussion on the inspiration and similar variants of the proposed extended object representation in Sections \ref{subsecOR} and \ref{subsecBPA}. (3) An overall procedure of the updated association decoding algorithm without limitation on the number and category of body-part as illustrated in Fig. \ref{network} and Alg. \ref{algDecoding}. (4) A seamless modification of BPJDet by using anchor-free detectors in Section \ref{AnchorFree}, as well as the new version BPJDetPlus with a further expansion of object representation for supporting many more multi-tasks in Section \ref{BPJDetPlus}. (5) New experiments on CrowdHuman and COCOHumanParts for body-head and body-parts joint detection of human in Sections \ref{subsecBHJD} and \ref{subsecBPJD}, respectively, and the newly built Animals5C for body-parts joint detection of five quadruped animals in Section \ref{subsecBHJDA}. Experimental results of more ablation and anchor-free BPJDet are also reported. (6) Two downstream applications including accurate crowd head detection and hand contact estimation in Sections \ref{subsecBHfACHD} and \ref{subsecBHfHCE} for explaining benefits of using the advanced body-head and body-hand association ability, respectively.

%%%%%%%%%%%%%%%%%%%%%%%%%%%%%%%%%%%%%%%%%%%%%%%%%%%%%%%%%%%%%%%
\vspace{-5pt}
\section{Related Work}\label{relwork}

% FastNFusion \cite{huang2021fast} fuses head features into the body features for occluded pedestrian detection. 

\subsection{Human Body and Part Detection}
We here only discuss emerging CNN-based detectors that achieve promising results over traditional approaches using hand-crafted features. For {\it human body} detection, it belongs to either general object detection containing the person category \cite{ren2015faster, lin2017feature, lin2017focal, redmon2016you, zhou2019objects, tian2019fcos, yolov5, yolov8} trained on common datasets like COCO \cite{lin2014microsoft}, or pedestrian detection \cite{ zhang2018occlusion, wang2018repulsion, chu2020detection, liu2019adaptive, xu2020beta} trained on specific benchmarks CityPersons \cite{zhang2017citypersons} and CrowdHuman \cite{shao2018crowdhuman}. A key challenge in human body detection is occlusion. Therefore, many researches propose customized loss functions \cite{zhang2018occlusion, wang2018repulsion}, improved NMS \cite{liu2019adaptive, chu2020detection} or tailored distribution model \cite{xu2020beta} for alleviating problems of crowded people detection. PedHunter \cite{chi2020pedhunter} designs a mask-guided module to leverage the head information to enhance the pedestrian feature. Currently, with the proposed burdensome yet powerful Detection Transformer (DETR) \cite{carion2020end}, some query-based crowd pedestrian detection methods \cite{lin2020detr, zhu2020deformable, sun2021sparse, zheng2022progressive} are dominant. On the other hand, detection of {\it body part} has also been intensively and extensively studied. Detection of face \cite{hu2017finding, deng2020retinaface}, head \cite{vu2015context, le2018key, peng2018detecting} and hand \cite{yang2020hier, narasimhaswamy2022whose} are all vigorous fields. Face detectors are usually based on well-designed networks and trained on dedicated datasets. While, head and hand detectors are rarely studied alone, and often used to facilitate downstream tasks. Precise design of body or part detector is beyond the scope of this paper. We focus on the joint detection of them.

\subsection{Body-Part Joint Detection}
We mainly pay attention to the joint detection of human body and one or more body parts like face, head and hands. First of all, {\it body-head} joint detection is the most popular couple, because the human head is a salient structural body part and plays a vital role in recognizing people, especially when occlusion occurs. For example, DA-RCNN \cite{zhang2019double} proposes to handle the crowd occlusion problem in human detection by capturing and cross-optimizing body and head parts in pairs with double anchor RPN. JointDet \cite{chi2020relational} presents a head-body relationship discriminating module to perform relational learning between heads and human bodies. Recently, BFJDet \cite{wan2021body} firstly investigates the performance of {\it body-face} joint detection, and proposes a bottom-up scheme that outputs body-face pairs for each pedestrian. It also adopts independent detection followed by pairwise association. For {\it body-hand} joint detection, \cite{zhou2018raising} devises heuristic strategies to match hand-raising gestures with body skeletons \cite{cao2017realtime} in classroom scenes. BodyHands \cite{narasimhaswamy2022whose} proposes a novel association network based on Mask R-CNN \cite{he2017mask} to jointly detect hands and the body location, and introduces a new corresponding hand-body association dataset. Besides, {\it body-parts} joint detection is also important for instance-level human-parts analysis. DID-Net \cite{li2019detector} adopts Faster R-CNN \cite{ren2015faster} with two carefully designed detectors for the human body and two body parts (hand and face) separately. Hier R-CNN \cite{yang2020hier} extends body parts up to six, and modifies Mask R-CNN \cite{he2017mask} with a new designed hierarchy branch inspired by FCOS \cite{tian2019fcos} for subordinate relationship learning of body and parts. Unlike all of them, our approach BPJDet is not restricted to one or more body parts, and can handle detection and association in an end-to-end way.

\subsection{Applications of Body-Part Association}
Superior body-part association is meaningful for boosting many downstream tasks. According to their dependence on body part associations, we can roughly divide them into two categories: {\it (1) low-level application:} such as person surveillance \cite{xiong2019vikingdet, shu2021head}, robust person ReID \cite{ding2020multi, somers2023body}, human parsing \cite{zhang2022aiparsing, yang2022quality, liu2022cdgnet} and human pose estimation \cite{dantone2014body, zhang2020learning, kreiss2021openpifpaf, cao2021openpose}. {\it (2) high-level understanding:} such as robot teleoperation \cite{gao2023parallel}, hand-related object contact \cite{brahmbhatt2020contactpose, narasimhaswamy2020detecting, shan2020understanding, muller2021self, chen2023detecting} and human-object interaction (HOI) \cite{li2020pastanet, wu2022mining, li2022hake, lim2023ernet}. Actually, there are essential differences between the two categories. The low-level tasks usually apply body-part associations to their methods for performance enhancing like robust person search by adding head \cite{shu2021head}, occluded person ReID relying on body part-based features \cite{somers2023body}, keypoints regressors using body parts dependence \cite{dantone2014body} and part affinity fields for learning to associate body parts \cite{cao2017realtime, kreiss2021openpifpaf, cao2021openpose}. Body-part association is an alternative for them. While, the high-level tasks are often inseparable from body-part associations. For example, human-machine systems should recognize the physical contact state of a body-part like hand \cite{brahmbhatt2020contactpose, gao2023parallel}. Human contact estimation \cite{narasimhaswamy2020detecting} and understanding \cite{shan2020understanding} may also need to analyze physical states of hands including their visual appearance and surrounding local context. In the well-defined field of HOI, body-part association is even more essential for understanding human activities \cite{li2020pastanet, li2022hake} and interactions \cite{wu2022mining, lim2023ernet}. To confirm the benefits of BPJDet, we choose to strengthen two basic representative tasks: low-level accurate crowd head detection and high-level hand contact estimation by establishing our proposed stronger body-part associations.

%%%%%%%%%%%%%%%%%%%%%%%%%%%%%%%%%%%%%%%%%%%%%%%%%%%%%%%%%%%%%%%
\vspace{-5pt}
\section{Preliminaries and Motivation}

In this section, we present and discuss works that largely motivate us to detect human body with regressing its parts jointly by proposing an extended object representation. 

\begin{figure}[!t]
	\centering
	\includegraphics[width=\columnwidth]{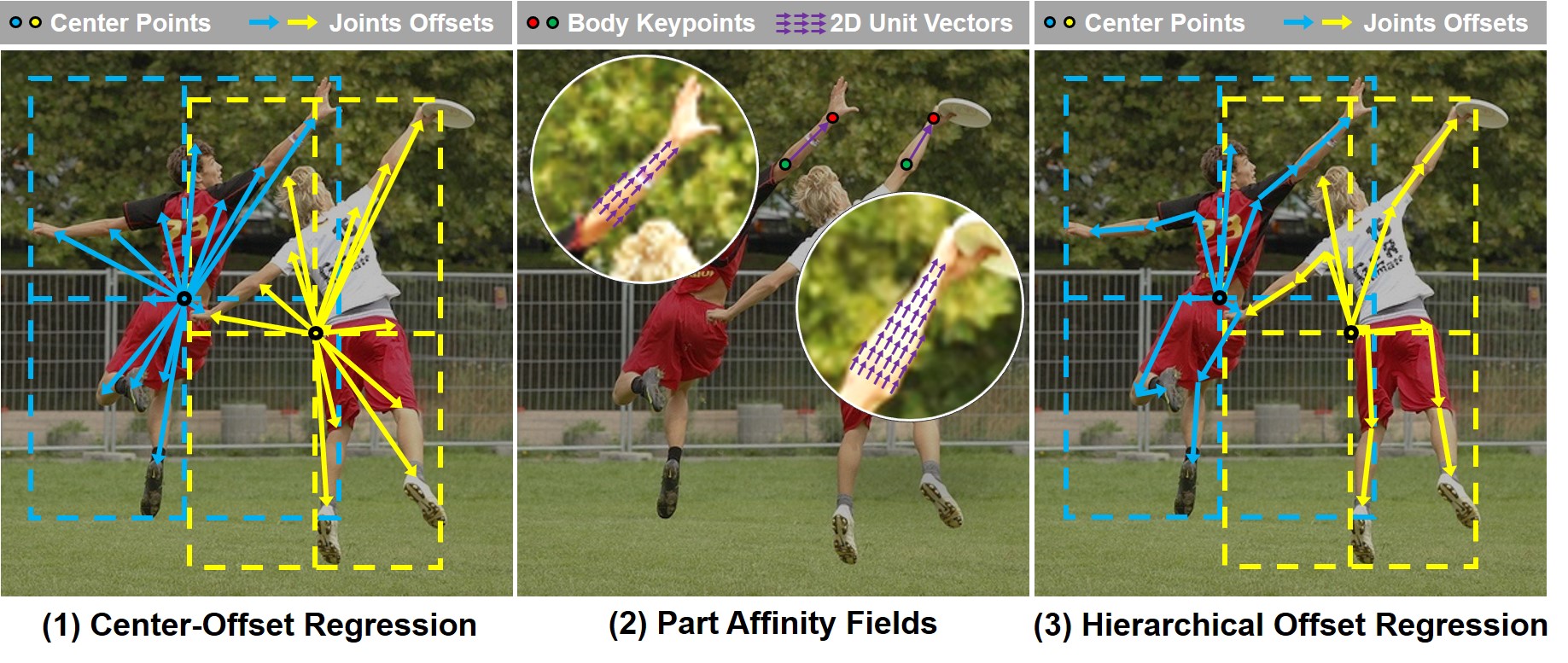}\\
	\vspace{-8pt}
	\caption{Illustrations of three popular human keypoints regression strategies: (1) {\it center-offset regression} used for direct structured pose representation,  (2) {\it part affinity fields} used for 2D unit direction vectors learning, (3) {\it hierarchical offset regression} used for hierarchical structured pose representation. We migrate them to the body-part association.}
	\label{bodypartAsso}
	\vspace{-10pt}
\end{figure}

\subsection{Object Representation}\label{subsecOR}
Most modern object detectors include three categories: {\it box-based} \cite{ren2015faster, redmon2016you, lin2017feature, lin2017focal, yolov5} using predefined anchors, {\it point-based} \cite{law2018cornernet, zhou2019objects, tian2019fcos, feng2021tood, li2022generalized, xu2022pp, yolov8} free of anchor boxes and {\it query-based} \cite{carion2020end, zhu2020deformable, sun2021sparse, zheng2022progressive, li2022dn} representing objects with a set of queries. Those {\it box-based} detectors heavily rely on handcrafted anchors that include parameters like number, size and aspect ratio. The object representation of them is rare and difficult to expand. Those {\it point-based} detectors directly predict object using keypoint or center point regression without needing of anchor boxes. This mechanism can eliminate the hyperparameters finetuning of anchors, achieve similar performance to anchor-based detectors with less computational cost, and keep better generalization ability. The last {\it query-based} DETRs abandon the traditional object representation and streamline the detection training pipeline by viewing it as a set prediction problem. However, their implicit encoder-decoder design does not align with our concept of extending representation. Instead, we propose a novel extended object representation by exploring both {\it box-based} (anchor-based) and {\it point-based} (anchor-free) detectors. 

Prior to us, some literatures have attempted to expand the point-based object representation, especially for human pose estimation that has an essential overlap with the object detection task. For example, CenterNet \cite{zhou2019objects} models objects using heatmap-based center points and subtly represents human poses as a 2K-dimensional property of the center point. Similarly, FCPose \cite{mao2021fcpose} adapts single-stage anchor-free FCOS \cite{tian2019fcos} with proposed dynamic filters to process person detections by predicting keypoint heatmaps and regressing offsets. Other single-stage pose estimators like SPM \cite{nie2019single}, AdaptivePose \cite{xiao2022adaptivepose} and YOLO-Pose \cite{maji2022yolo} also learn to regress keypoints as series of joint offsets. Recently, ED-Pose \cite{yang2023explicit} presents an explicit box detection framework which unifies the contextual learning between global human-level and local keypoint-level information based on DN-DETR \cite{li2022dn}. It re-considers human pose estimation task as two explicit box detection processes with a unified representation and regression supervision, which much resembles our joint body (global) and part (local) detection design. % In this way, pose estimation is regarded as a keypoint box detection problem to learn both box positions and contents for each keypoint

\subsection{Body-Part Association}\label{subsecBPA}
Currently, most body-part joint detection methods \cite{li2019detector, yang2020hier, chi2020relational, wan2021body, narasimhaswamy2022whose} are based on two separate stages including detection and post-matching. The inherent drawbacks of post-matching for body-part association are often unavoidable. For example, once the true main body of one detected body-part is undetected or improperly boxed, it will be a high probability event by mistakenly allocating this body-part to one person bounding box it just falls within, even if the error is quite obvious or unreasonable. In Section \ref{QVC}, we give some ordinary failure cases of body-hand in Fig. \ref{BPJDetHand} and body-parts in Fig. \ref{BPJDetParts} to explicitly reflect their defects. To alleviate this problem, we now have at least three strategies originally designed for human pose estimation to complete a single-stage body-part association. They are (1) {\it center-offset regression} as in CenterNet \cite{zhou2019objects} and FCOS \cite{tian2019fcos}, (2) {\it part affinity fields} as in OpenPose \cite{cao2017realtime, cao2021openpose} and OpenPifPaf \cite{kreiss2021openpifpaf}, and (3) {\it hierarchical offset regression} as in SPM \cite{nie2019single}. We illustrate their discrepancies in Fig. \ref{bodypartAsso}. These correlation methods have greatly inspired our work. We apply the first intuitive yet still sufficiently effective center-offset regression strategy to validate the desirability to extend this single-stage association to traditional  object representation.

%%%%%%%%%%%%%%%%%%%%%%%%%%%%%%%%%%%%%%%%%%%%%%%%%%%%%%%%%%%%%%%
\vspace{-5pt}
\section{Our Method}

\begin{figure*}[!t]
	\centering
	\includegraphics[width=0.9\textwidth]{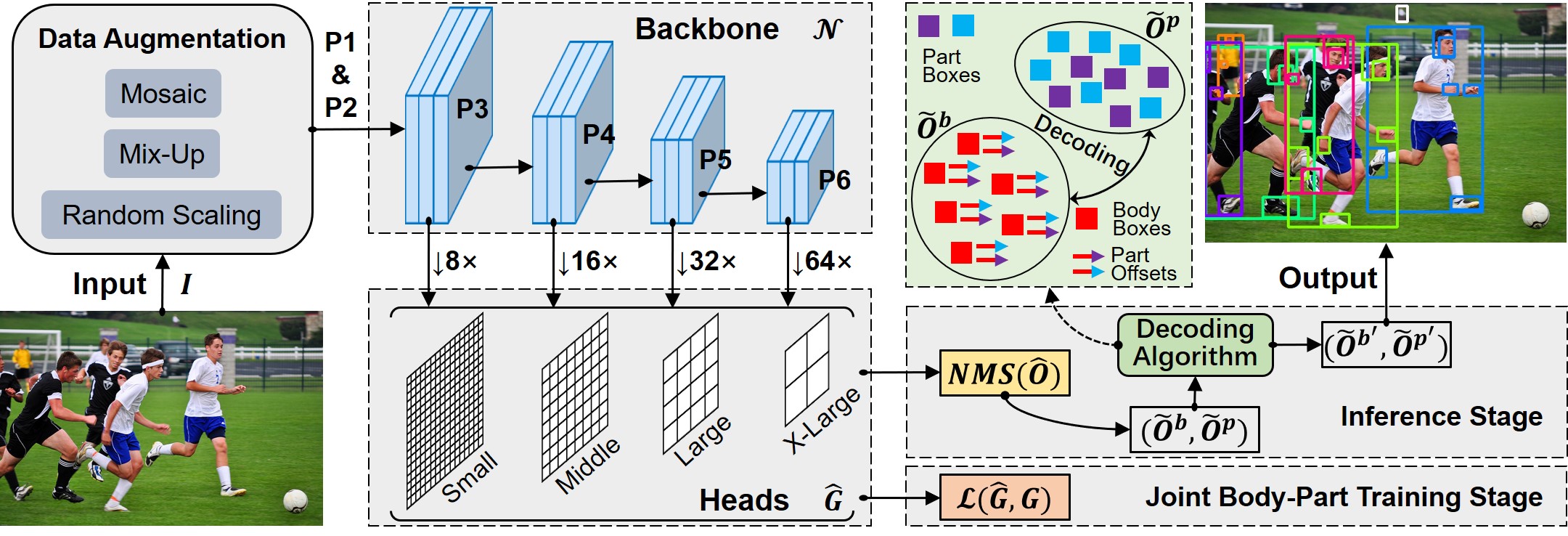}\\
	\vspace{-8pt}
	\caption{Our BPJDet adopts YOLOv5 as the backbone $\mathcal{N}$ to extract features and predict grids $\mathit{\widehat{G}}$ from one augmented input image $\mathbf{I}$. During training, target grids $\mathit{G}$ are used to supervise the elaborately designed multi-loss function $\mathcal{L}$. In inference stage, NMS and association decoding algorithm are sequentially applied on predicted objects $\widehat{\mathbf{O}}$ to obtain final human body boxes set $\widetilde{\mathbf{O}}^{b'}$ and related body parts set $\widetilde{\mathbf{O}}^{p'}$.}
	\label{network}
	\vspace{-10pt}
\end{figure*}

\subsection{Extended Object Representation}

In our proposed BPJDet, we train a dense single-stage anchor-based detector to directly predict a set of objects $\{\widehat{\mathcal{O}}\!\in\!\widehat{\mathbf{O}}\!\parallel\!\widehat{\mathcal{O}}\!=\!cat(\widehat{\mathcal{O}}^{box},\widehat{\mathcal{O}}^{dis}), \widehat{\mathbf{O}}\!=\!\widehat{\mathbf{O}}^b \cup \widehat{\mathbf{O}}^p\}$, which contains human body set $\widehat{\mathbf{O}}^b$ and body-part set $\widehat{\mathbf{O}}^p$ concurrently. A typical extended object prediction $\widehat{\mathcal{O}}$ is concatenated of the bounding box $\widehat{\mathcal{O}}^{box}$ and corresponding center point displacement $\widehat{\mathcal{O}}^{dis}$. 
\begin{equation}%\small
\begin{aligned}
	\widehat{\mathcal{O}}& = cat(\widehat{\mathcal{O}}^{box},\widehat{\mathcal{O}}^{dis}) = (\hat{o}, \mathbf{\hat{b}}, \mathbf{\hat{c}}, \mathbf{\hat{d}}) \\
	&= (\hat{o}, \hat{b}_x, \hat{b}_y, \hat{b}_w, \hat{b}_h, \hat{c}_1, ..., \hat{c}_{k+1}, \hat{d}_{x_1}, \hat{d}_{y_1}, ..., \hat{d}_{x_k}, \hat{d}_{y_k}) ~
	\label{oneprediction}
\end{aligned}
\end{equation}
It locates an object with a tight bounding box $\mathbf{\hat{b}}\!=\!(\hat{b}_x, \hat{b}_y, \hat{b}_w, \hat{b}_h)$ where coordinates $(\hat{b}_x, \hat{b}_y)$ are the center position, $\hat{b}_w$ and $\hat{b}_h$ are the width and height of $\mathbf{\hat{b}}$, respectively. It also records the relative displacement $\mathbf{\hat{d}}\!=\!(\hat{d}_{x_i}, \hat{d}_{y_i})|^k_{i=1}$ of center point of body-part ($\widehat{\mathcal{O}}\!\in\!\widehat{\mathbf{O}}^b$) or affiliated body ($\widehat{\mathcal{O}}\!\in\!\widehat{\mathbf{O}}^p$) to each $\mathbf{\hat{b}}$. Here, $k$ is the body-part number. Considering that body-part may be one or more components (e.g., two both hands with $k\!=\!2$, and the COCOHumanParts with $k\!=\!6$), we allow $\mathbf{\hat{d}}$ to be compatible with dynamic 2D offsets. The $\hat{o}$ and $\mathbf{\hat{c}}$ are predicted objectness and classification scores, respectively. For these regressed offsets, we will explain how to decode them for the body-part association with fusing detected body-part set $\widehat{\mathbf{O}}^p$ in Section \ref{inference}.

Intuitively, we can benefit a lot from this extended representation. On one hand, an appropriate large body bounding box $\widehat{\mathcal{O}}^{box}$ possesses both strong local characteristics and weak global features (such as surrounding background and anatomical position) for its body part $\widehat{\mathcal{O}}^{dis}$ offset regression. This enables the network to learn their intrinsic relationships. On the other hand, compared to methods \cite{yang2020hier, chi2020relational, wan2021body, narasimhaswamy2022whose} training multiple subnetworks or stages, mixing $\widehat{\mathcal{O}}^{box}$ and $\widehat{\mathcal{O}}^{dis}$ up for synchronous learning can be leveraged directly in BPJDet without the need of complicated post-processing. By designing a one-stage network that uses shared heads to jointly predict $\widehat{\mathcal{O}}^{box}$ and $\widehat{\mathcal{O}}^{dis}$, our approach can achieve high accuracy with much less computational burden during training and inference.

%%%%%%%%%%%%
\subsection{Overall Network Architecture}\label{netarch}

Our network structure is shown in Fig.~\ref{network}. We choose the recent cost-effective one-stage YOLOv5 \cite{yolov5} as the basic backbone $\mathcal{N}$. Specifically, following YOLOv5, we feed one RGB image $\mathbf{I}\!\in\!\mathbb{R}^{h\times w\times 3}$ as the input to $\mathcal{N}$, keep its beneficial data augmentation strategies (e.g., Mosaic and MixUp), and output four grids $\widehat{\mathit{G}}\!=\!\{\widehat{\mathcal{G}}^s\!\parallel\!s\!\in\!\{8,16,32,64\}\}$ from four multi-scale heads. Each grid $\widehat{\mathcal{G}}\!\in\!\mathbb{R}^{A_a\times A_o\times \frac{h}{s}\times \frac{w}{s}}$ contains dense object outputs $\widehat{\mathbf{O}}$ produced from $A_a$ anchor channels and $A_o$ output channels. Supposing that one target object $\mathcal{O}$ is centered at $(b_x, b_y)$ in the feature map $\mathbf{F}^s$, the corresponding grid $\widehat{\mathcal{G}}^s$ at cell $({b_x}/{s}, {b_y}/{s})$ should be highly confident. When having defined $A_a$ anchor boxes $\mathcal{B}^s\!=\!\{(B^w_i,  B^h_i)\vert^{A_a}_i\}$ for the grid $\widehat{\mathcal{G}}^s$, we will generate $A_a$ anchor channels at each cell $({b_x}/{s}, {b_y}/{s})$. Furthermore, to obtain robust capability, YOLOv5 allows detection redundancy of multiple objects and four surrounding cells matching for each cell. This redundancy makes sense to the detection of body or part position $\mathcal{O}^{box}$, but it is not completely facilitative to the regression of offsets $\mathcal{O}^{dis}$. We interpret this in Section \ref{loss}.

\begin{figure}[!t]
	\centering
	\includegraphics[width=\columnwidth]{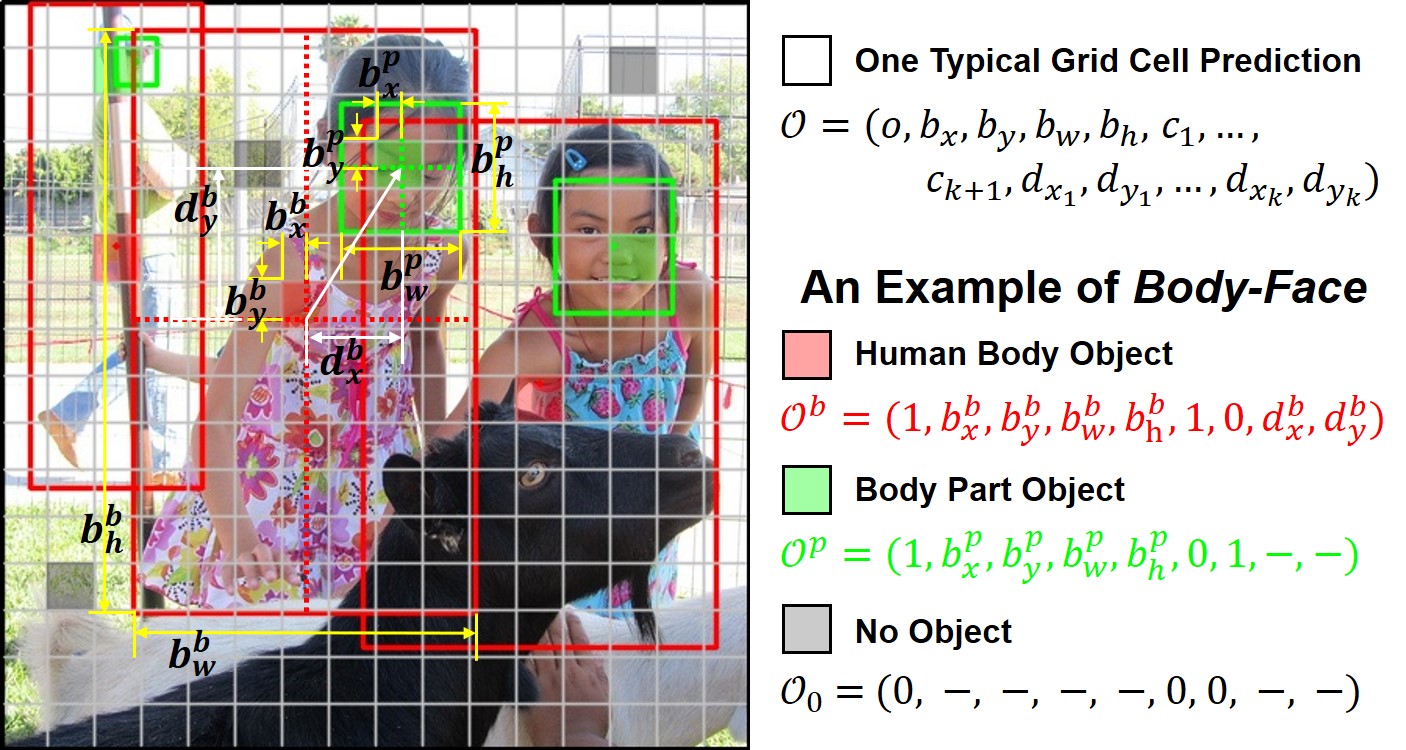}\\
	\vspace{-8pt}
	\caption{Examples for grid cell predictions with human body objects in red color and body part objects (e.g., face) in green color. The ''--'' means not used when calculating training losses.}
	\label{gridcell}
	\vspace{-10pt}
\end{figure}

Then, we explain the arrangement of one prediction $\widehat{\mathcal{O}}$ from $A_o$ output channels of $\widehat{\mathcal{G}}^s_{i,x,y}$ which is related to $i$-th anchor box at grid cell $(x,y)$. As shown in the example of grid cells in Fig.~\ref{gridcell}, one typical predicted $\widehat{\mathcal{O}}$ embedding consists of four parts: the objectness or probability $\hat{o}$ that an object exists, the bounding box candidate $\mathbf{\hat{b}}'=({\hat{b}_x}', {\hat{b}_y}', {\hat{b}_w}', {\hat{b}_h}')$, the object classification scores $\mathbf{\hat{c}'}\!=\!(\hat{c'}_1, ..,\hat{c'}_{k+1})$, and the body-part offsets candidate $\mathbf{\hat{d}}'\!=\!(\hat{d}'_{x_i}, \hat{d}'_{y_i})|^k_{i=1}$. Thus, $A_o\!=\!3k+6$. To transform the candidate $\mathbf{\hat{b}}'$ into coordinates $\mathbf{\hat{b}}$ relative to the grid cell $\widehat{\mathcal{G}}^s_{i,x,y}$, we apply conversions as below:
\begin{equation}%\small
\begin{aligned}
	&{\hat{b}_x}=2\phi({\hat{b}_x}')-0.5, \qquad\enspace\enspace\; {\hat{b}_y}=2\phi({\hat{b}_y}')-0.5 ~\\
	&{\hat{b}_w}=[2\phi({\hat{b}_w}')]^2{B^w_i}/{s}, \qquad {\hat{b}_h}=[2\phi({\hat{b}_h}')]^2{B^h_i}/{s} ~
\end{aligned}
\end{equation}
where $\phi$ is the sigmoid function that limits model predictions in the range of $(0,1)$. Similarly, this detection strategy can be extended to the offset of body-part. A body-part’s intermediate offsets $\mathbf{\hat{d}}'$ are predicted in the grid coordinates and relative to the grid cell origin $(x,y)$ using:
\begin{equation}%\small
	{\hat{d}_{x_i}}\!=\![4\phi({\hat{d}_{x_i}}')-2]{B^w_i}/{s}, \:\:\:\:\:\:\:\: \\
	{\hat{d}_{y_i}}\!=\![4\phi({\hat{d}_{y_i}}')-2]{B^h_i}/{s} ~
\end{equation}
In the way, $\hat{d}_{x_i}$ and $\hat{d}_{y_i}$ are constrained to ranges $\pm2{B^w_i}/{s}$ and $\pm2{B^h_i}/{s}$, respectively. To learn $\mathbf{\hat{b}}$ and $\mathbf{\hat{d}}$, losses are applied in the grid space. Sample targets of $\mathbf{b}$ and $\mathbf{d}$ are shown in Fig.~\ref{gridcell}.

%%%%%%%%%%%%
\subsection{Multi-Loss Functions}\label{loss}

For a set of predicted grids $\mathit{\widehat{G}}$, we firstly build target grids set $\mathit{G}$ following formats introduced in Section \ref{netarch}. Then, we mainly compute following four loss components:
\begin{align}%\footnotesie %\small
	& \mathcal{L}_{box}=\sum\nolimits_s\frac{1}{\|\mathcal{G}^s\|}\sum\nolimits^{\|\mathcal{G}^s\|}_{i=1}[1-\mathsf{CIoU}(\mathbf{\hat{b}_i}, \mathbf{b_i})] \\
	& \mathcal{L}_{obj}=\sum\nolimits_s\frac{w_s}{\|\mathcal{G}^s\|}\sum\nolimits^{\|\mathcal{G}^s\|}_{i=1}\mathsf{BCE}(\hat{o}, o\cdot\mathsf{CIoU}(\mathbf{\hat{b}_i}, \mathbf{b_i})) \\
	& \mathcal{L}_{cls}=\sum\nolimits_s\frac{1}{\|\mathcal{G}^s\|}\sum\nolimits^{\|\mathcal{G}^s\|}_{i=1}\mathsf{BCE}(\hat{c}, c) \\
	& \mathcal{L}_{bpd}=\sum\nolimits_s\frac{1}{\|\mathcal{G}^s\|}\sum\nolimits^{\|\mathcal{G}^s\|}_{i=1}\sum\nolimits^{k}_{1}\varphi(v>0)\| \mathbf{\hat{d}'_i}-\mathbf{d'_i} \|_2 ~
\end{align}

For the bounding box regression loss $\mathcal{L}_{box}$, we adopt the complete intersection over union (CIoU) across four grids $\mathcal{G}^s$. $\mathsf{BCE}$ in both the objectness loss $\mathcal{L}_{obj}$ and classification loss $\mathcal{L}_{cls}$ is the binary cross-entropy. The multiplier $o$ in $\mathcal{L}_{obj}$ is used for penalizing candidates without hitting target grid cells ($o\!=\!0$), and encouraging candidates around target anchor ground-truths ($o\!=\!1$). The coefficient $w_s$ is a balance weight for different grid levels in YOLOv5. For the body-part displacement loss $\mathcal{L}_{bpd}$, we utilize the mean squared error (MSE) to measure offset outputs $\mathbf{\hat{d}'}$ and normalized targets $\mathbf{d'}$. Before that, we apply a filter $\varphi(\cdot)$ with the visibility label of body-part to remove out those false-positive offset predictions from $\widehat{\mathcal{O}}$. Finally, we calculate the total training loss $\mathcal{L}=(\mathit{\widehat{G}},\mathit{G})$ as follows:
\begin{equation}%\small
	\mathcal{L}=N_{bs}(\alpha\mathcal{L}_{box} + \beta\mathcal{L}_{obj} + \gamma\mathcal{L}_{cls} + \lambda\mathcal{L}_{bpd}) ~
	\label{losstotal}
\end{equation}
where $N_{bs}$ is the batch size. The $\alpha$, $\beta$, $\gamma$ and $\lambda$ are weights of losses $\mathcal{L}_{box}$, $\mathcal{L}_{obj}$, $\mathcal{L}_{cls}$ and $\mathcal{L}_{bpd}$, respectively.

%%%%%%%%%%%%

\begin{algorithm}[t]
\caption{Overall Procedure of Association Decoding.}
\begin{algorithmic}\footnotesize
	\STATE $\bullet$ \textbf{Input:} Body and part object sets ($\widetilde{\mathbf{O}}^{b}$, $\widetilde{\mathbf{O}}^{p}$), inner IoU threshold $\tau^{inner}_{iou}$, body-part catgegories and numbers $k$.
	\STATE $\bullet$ \textbf{Output:} Body and parts boxes set pairs ($\widetilde{\mathbf{O}}^{b'}$, $\widetilde{\mathbf{O}}^{p'}$).

	\STATE Initialize $\widetilde{\mathbf{O}}^{b'} = \widetilde{\mathbf{O}}^{p'} = \emptyset$;
	\STATE $n_b \leftarrow \widetilde{\mathbf{O}}^{b}.size(0)$; \hfill\textcolor{olive}{// body objects number}
	\STATE $\widetilde{\mathbf{P}} \leftarrow \widetilde{\mathbf{O}}^{b}[:, -2k:].reshape(n_b, k, 2)$; \hfill\textcolor{olive}{// part center points by offset}
	\STATE $\widetilde{\mathbf{P}} \leftarrow \textsc{cat}(\widetilde{\mathbf{P}}, np.zeros(n_b, k, 5))$; \hfill\textcolor{olive}{// extend it to save associated box}
	
	\STATE \textbf{for} {$(\widetilde{\mathcal{O}}^{p}_{box}, \widetilde{\mathcal{O}}^{p}_{conf}, \widetilde{\mathcal{O}}^{p}_{cls})$ \textbf{in} $\widetilde{\mathbf{O}}^{p}$} \textbf{do}
   		\STATE \hspace{0.4cm} $(x^p_1, y^p_1, x^p_2, y^p_2) \leftarrow \widetilde{\mathcal{O}}^{p}_{box}$;
   		\STATE \hspace{0.4cm} $(c^p_x, c^p_y) \leftarrow ((x^p_1+x^p_2)/2,  (y^p_1+y^p_2)/2)$; \hfill\textcolor{olive}{// part center point}
   		\STATE \hspace{0.4cm} $\widetilde{\mathbf{P}}_{cls} \leftarrow \widetilde{\mathbf{P}}[:, \widetilde{\mathcal{O}}^{p}_{cls}]$; \hfill\textcolor{olive}{// select specific parts}
   		\STATE \hspace{0.4cm} $\widetilde{\mathbf{P}}_{dist} \leftarrow \|\widetilde{\mathbf{P}}_{cls}[:,:2]-(c^p_x, c^p_y)\|$; \hfill\textcolor{olive}{// calculate L2 distance}
   		\STATE \hspace{0.4cm} $index \leftarrow \textsc{argmin}(\widetilde{\mathbf{P}}_{dist})$;
   		\STATE \hspace{0.4cm} $\widetilde{\mathcal{O}}^{b}_{box} \leftarrow \widetilde{\mathbf{O}}^{b}[index,:4]$; \hfill\textcolor{olive}{// fetch nearest body box}
   		\STATE \hspace{0.4cm} $iou \leftarrow \textsc{inner\_iou}(\widetilde{\mathcal{O}}^{b}_{box}, \widetilde{\mathcal{O}}^{p}_{box})$;
		\STATE \hspace{0.4cm} \textbf{if} {$(\widetilde{\mathcal{O}}^{p}_{conf}>\widetilde{\mathbf{P}}_{cls}[index, 2])$ \textbf{and} $(iou>\tau^{inner}_{iou})$} \textbf{then}
			\STATE \hspace{0.8cm} $\widetilde{\mathbf{P}}[index, \widetilde{\mathcal{O}}^{p}_{cls}] \leftarrow [c^p_x, c^p_y, \widetilde{\mathcal{O}}^{p}_{conf}, x^p_1, y^p_1, x^p_2, y^p_2]$; \hfill\textcolor{olive}{// update}
		\STATE \hspace{0.4cm} \textbf{end}
	\STATE \textbf{end}
	
	\STATE $\widetilde{\mathbf{O}}^{b'} \leftarrow \widetilde{\mathbf{O}}^{b}[:,:4]$; \hfill\textcolor{olive}{// body objects boxes with shape ($n_b$, 4)}
	\STATE $\widetilde{\mathbf{O}}^{p'} \leftarrow \widetilde{\mathbf{P}}[:,:,-4:]$; \hfill\textcolor{olive}{// parts objects boxes with shape ($n_b$, k, 4)}
	
	\STATE \textbf{return} $(\widetilde{\mathbf{O}}^{b'}, \widetilde{\mathbf{O}}^{p'})$;
\end{algorithmic}
\label{algDecoding}
\end{algorithm}
%\vspace{-10pt}

\vspace{-5pt}
\subsection{Association Decoding}\label{inference}

In inference, we process predicted objects set $\widehat{\mathbf{O}}$ to get final results. First of all, we apply the conventional Non-Maximum Suppression (NMS) to filter out false-positive and redundant bounding boxes of body and part objects:
\begin{align}
	& \widehat{\mathbf{O}}^{b'}\!= \mathsf{NMS}(\widehat{\mathbf{O}}^b, \tau^b_{conf}, \tau^b_{iou}) \\  % \:\:\:\: \\
	& \widehat{\mathbf{O}}^{p'}\!= \mathsf{NMS}(\widehat{\mathbf{O}}^p, \tau^p_{conf}, \tau^p_{iou})~
\end{align}
where $\tau_{conf}$ and $\tau_{iou}$ are thresholds for object confidence and IoU overlap, respectively. Considering different recognition difficulties and overlapping levels, we define distinct thresholds $\tau^b$ and $\tau^p$ for filtering body and part boxes. We fetch confidence of each predicted object $\widehat{\mathcal{O}}$ by $\hat{o}\cdot\hat{c}_i$, where $i\!=\!1$ for body object and $i\!>\!1$ for part object.  Then, we rescale $\widehat{\mathcal{O}}^{box'}$ and $\widehat{\mathcal{O}}^{dis'}$ in $\widehat{\mathbf{O}}^{b'}$ and $\widehat{\mathbf{O}}^{p'}$ to obtain real $\widetilde{\mathbf{O}}^{b}$ and $\widetilde{\mathbf{O}}^{p}$ by the following transformations:
\begin{align}
	& \widetilde{\mathcal{O}}^{box}\!=\!s\cdot[\widehat{\mathcal{O}}^{box'}+(x_o, y_o, 0, 0)] \\ % \:\:\:\: \\
	& \widetilde{\mathcal{O}}^{dis}\!=\!s\cdot[\widehat{\mathcal{O}}^{dis'}+(x_o, y_o)] ~
\end{align}
where $x_o$ and $y_o$ are offsets from grid cell centers. The scalar $s$ maps the down-sampled size of box and offset back to the original image shape.

Finally, we update associated body parts of each left body object in $\widetilde{\mathbf{O}}^{b}$ by fusing its regressed center point offsets ($\widetilde{\mathcal{O}}^{dis}\!\in\!\widehat{\mathbf{O}}^{b'}$) with the remaining candidate part objects $\widetilde{\mathbf{O}}^{p}$. Specifically, we search each box $\widetilde{\mathcal{O}}^{box}$ in part objects with its nearest body-part offset $\widetilde{\mathcal{O}}^{dis}$ belonging to body object. The part box that has a large inner IoU ($>\tau^{inner}_{iou}$) with the body box will be selected. Inner IoU here represents the ratio of intersection area on smaller body-part bounding box instead of the larger union of body-part and human body.
\begin{equation}\small
\begin{aligned}
	& \quad IoU = \frac{BOX_{body} \bigcap BOX_{part}}{BOX_{body} \bigcup BOX_{part}} ~\\
	& \!\!\!\!\!\! \!\!\!\!\!\! Inner\;IoU = \frac{BOX_{body} \bigcap BOX_{part}}{BOX_{part}} ~
	\label{innerIoU}
\end{aligned}
\end{equation}
We here naturally assume that the smaller body-part box should be completely inside its associated larger human body box if using ground-truth labels. During the prediction stage, detected boxes do not always tightly surround the objects. It may be too large or too small. We cannot simply set $\tau^{inner}_{iou}\!=\!1$. In contrast, a very small $\tau^{inner}_{iou}$ will obviously lead to many wrong body part pair candidates. Therefore, we should assign a reasonable $\tau^{inner}_{iou}$ to keep a balance between the quantity and quality of matching pairs. This experimental discussion is presented in Section \ref{tauAS}. We summarize the overall decoding process in Alg. \ref{algDecoding}. We report all evaluation results on the final updated body and parts boxes set pairs $(\widetilde{\mathbf{O}}^{b'}, \widetilde{\mathbf{O}}^{p'})$.

%%%%%%%%%%%%
\subsection{BPJDet using Anchor-Free Detectors}\label{AnchorFree}

To reveal the universality of our proposed extended representation, we here interpret how to integrate it with advanced anchor-free detectors YOLOv5u \cite{yolov5} and YOLOv8 \cite{yolov8}. YOLOv8 is the latest iteration of YOLO series, which employs state-of-the-art backbone \cite{feng2021tood, li2022generalized} and neck architectures. It adopts an objectness-free split head, which contributes to better accuracy and a more efficient detection process. YOLOv5u modernizes YOLOv5 by originating from its foundational architecture and adopting an anchor-free head as in YOLOv8. These similarities allow us to extensively reuse the detailed designs in the previous subsections, with a key focus on representation adjustment.

Specifically, in the anchor-free paradigm, detection is formulated as a dense inference of every pixel in feature maps. For each position $p_i\!=\!(x_i, y_i)$ in $\mathbf{F}^s$, the prediction head regresses a 4D vector $(l_i, t_i, r_i, b_i)$, which represents the relative offsets from the four sides (left, top, right and bottom) of a bounding box anchored in $p_i$. We reformulate the anchor-based extended representation $\widehat{\mathcal{O}}$ in Eqn. \ref{oneprediction} into the anchor-free one $\widehat{\mathcal{O}}_u$ as below.
\begin{equation}%\small
\begin{aligned}
	\widehat{\mathcal{O}}_u& = cat(\widehat{\mathcal{O}}_u^{box},\widehat{\mathcal{O}}_u^{dis}) = (\mathbf{\hat{b}}_u, \mathbf{\hat{c}}_u, \mathbf{\hat{d}}_u) \\
	&= (\hat{b}_l, \hat{b}_t, \hat{b}_r, \hat{b}_b, \hat{c}_1, ..., \hat{c}_{k+1}, \hat{d}_{x_1}, \hat{d}_{y_1}, ..., \hat{d}_{x_k}, \hat{d}_{y_k}) ~
	\label{onepredictionfree}
\end{aligned}
\end{equation}
where the objectness score $\hat{o}$ is omitted. The meaning of $\mathbf{\hat{c}}_u$ and $\mathbf{\hat{d}}_u$ are totally the same as $\mathbf{\hat{c}}$ and $\mathbf{\hat{d}}$, respectively. The bounding box $\mathbf{\hat{b}}_u\!=\!(\hat{b}_l, \hat{b}_t, \hat{b}_r, \hat{b}_b)$ is the distinctive design for {\it point-based} detectors. In both YOLOv5u and YOLOv8, they adopt the same task-aligned assigner as in TOOD \cite{feng2021tood} to learn the probabilities of values around the continuous locations of target bounding boxes $\widehat{\mathcal{O}}_u^{box}$, which includes a CIoU loss $\mathcal{L}_{u_{box}}$ and a Distribution Focal Loss (DFL) \cite{li2022generalized} $\mathcal{L}_{u_{dfl}}$. Please refer \cite{feng2021tood} and \cite{li2022generalized} for the explicit definition of these two losses. The classification loss $\mathcal{L}_{u_{cls}}$ uses the binary cross-entropy. For the body-part displacement loss $\mathcal{L}_{u_{bpd}}$, we synchronously generate target GTs in one identical box assigner, and optimize corresponding outputs $\mathbf{\hat{d}'}_u$ using MSE. The final total training loss $\mathcal{L}_u$ is as follows:
\begin{equation}%\small
	\mathcal{L}_u=N_{bs}(\alpha_u\mathcal{L}_{u_{box}} + \beta_u\mathcal{L}_{u_{dfl}} + \gamma_u\mathcal{L}_{u_{cls}} + \lambda_u\mathcal{L}_{u_{bpd}}) ~
	\label{losstotalPlus}
\end{equation}
which is an updating of $\mathcal{L}$ in Eqn. \ref{losstotal}. $N_{bs}$ is the batch size. The $\alpha_u$, $\beta_u$, $\gamma_u$ and $\lambda_u$ are weights of each loss. In inference, the association decoding part keeps unchanged.

%%%%%%%%%%%%
\subsection{BPJDetPlus: Further Extended Representation}\label{BPJDetPlus}

For supporting the downstream application hand contact estimation, we build a new architecture by extending the proposed object representation with additional spaces to estimate the contact state of a detected hand. We rename this new framework as BPJDetPlus. 

Specifically, we keep extending the object representation $\widehat{\mathcal{O}}$ with adding $\widehat{\mathcal{O}}^{cts}$ that represents four contact states of two hands . The new typical prediction $\widehat{\mathcal{O}'}$ inherited from $\widehat{\mathcal{O}}$ in Eqn. \ref{oneprediction} is constituted as below:
\begin{equation}%\small
\begin{aligned}
	\widehat{\mathcal{O}'}& = cat(\widehat{\mathcal{O}}^{box},\widehat{\mathcal{O}}^{dis}, \widehat{\mathcal{O}}^{cts}) = (\hat{o}, \mathbf{\hat{b}}, \mathbf{\hat{c}}, \mathbf{\hat{d}}, \mathbf{\hat{h}}) \\
	&= (\hat{o}, \hat{b}_x, \hat{b}_y, \hat{b}_w, \hat{b}_h, \hat{c}_1, ..., \hat{c}_{k+1}, \hat{d}_{x_1}, \hat{d}_{y_1}, ..., \\
	& \quad \hat{d}_{x_k}, \hat{d}_{y_k}, \hat{h}^1_{s_1}, \hat{h}^1_{s_2}, \hat{h}^1_{s_3}, \hat{h}^1_{s_4}, \hat{h}^2_{s_1}, \hat{h}^2_{s_2}, \hat{h}^2_{s_3}, \hat{h}^2_{s_4}) ~
	\label{onenewprediction}
\end{aligned}
\end{equation}
where $\mathbf{\hat{h}}$ contains hand contact states. The $\hat{h}^1_{s_j}$ or $\hat{h}^2_{s_j}$ is a specific contact state $s_j$ ($j \in \{1,2,3,4\}$) of one hand part. Following \cite{narasimhaswamy2020detecting}, we define a new hand contact states loss $\mathcal{L}_{cts}$ to be the sum of four independent binary cross-entropy losses corresponding to four possible categories. 
 \begin{equation}%\small
	\mathcal{L}_{cts}=\sum\nolimits_s\frac{1}{\|\mathcal{G}^s\|}\sum\nolimits^{\|\mathcal{G}^s\|}_{i=1}\sum\nolimits^{4}_{j=1}\mathsf{BCE}(\hat{h}_{s_j}, h_{s_j}) \\
\end{equation}
where $\hat{h}_{s_j}$ is the predicted hand contact state ranging from 0 to 1, and $h_{s_j}$ is the ground-truth state being 0, 1 or 2 meaning {\it No, Yes} or {\it Unsure}, respectively. We do not penalize $h_{s_j}$ with {\it Unsure} state. The final loss of BPJDetPlus is updated as:
\begin{equation}%\small
	\mathcal{L}'=N_{bs}(\alpha\mathcal{L}_{box} + \beta\mathcal{L}_{obj} + \gamma\mathcal{L}_{cls} + \lambda\mathcal{L}_{bpd} + \mu\mathcal{L}_{cts}) ~
	\label{losstotalPlus}
\end{equation}
which is an extension of the multi-loss function $\mathcal{L}$ in Eqn. \ref{losstotal}. $N_{bs}$ is the batch size. The loss weight $\mu$ of $\mathcal{L}_{cts}$ is empirically set as 0.01 by online searching similar to $\lambda$ in Section \ref{lambdaASs}. We train BPJDetPlus on the train-set of ContactHands \cite{narasimhaswamy2020detecting} and evaluate its performance on the test-set. In inference, the final predicted hand contact state score is a weighted sum of two probabilities embedded in the hand and its associated body instance. More results are in Section \ref{subsecBHfHCE}.

%%%%%%%%%%%%%%%%%%%%%%%%%%%%%%%%%%%%%%%%%%%%%%%%%%%%%%%%%%%%%%%
\section{Experimental Results}

%%%%%%%%%%%%%%%%%%%%%%%%

\subsection{Experimental Settings}

\subsubsection{Datasets}\label{datasetsIntro}
{\bf Body-Part Joint Detection:} We firstly expect to evaluate the association quality of proposed BPJDet, while maintaining high object detection accuracy. We choose four wildly used benchmarks including CityPersons \cite{zhang2017citypersons}, CrowdHuman \cite{shao2018crowdhuman}, BodyHands \cite{narasimhaswamy2022whose} and COCOHumanParts \cite{yang2020hier}. The former two are for pedestrian detection tasks. In CityPersons, it has 2,975 and 500 images for training and validation, respectively. It only provides box labels of pedestrians. In CrowdHuman, there are 15,000 images for training and 4,375 images for validation. It provides box labels of both body and head. Following BFJDet \cite{wan2021body}, we use its re-annotated box labels of visible faces for conducting corresponding {\it body-face} joint detection experiments. Besides, we supply the {\it body-head} joint detection experiment on CrowdHuman as a strong baseline. The third dataset BodyHands is for hand-body association tasks. It has 18,861 and 1,629 images in train-set and test-set with annotations for hand and body locations and correspondences. We implement {\it body-hand} joint detection task in it for comparing. The last dataset COCOHumanParts contains 66,808 images with 64,115 in train-set and 2,693 in val-set. It has inherited bounding-box of person category from official COCO, and labeled the locations of six body-parts (face, head, right-hand/left-hand and right-foot/left-foot) in each instance if it is visible. We execute {\it body-parts} joint detection task in it for comparing with Hier R-CNN \cite{yang2020hier}. The statistics of these four datasets are summarized in Table \ref{dataStat}.

Besides, to verify the generality of BPJDet about body-part association, we reconstruct a dataset Animals5C with five kinds of quadruped animals (e.g., dog, cat, sheep, horse and cow) based on AnimalPose \cite{cao2019cross} and AP-10K \cite{yu2021ap}. We adopt their bounding box labels of the animal body, and generate boxes of five parts (e.g., head and four feet) using keypoints. Finally, we obtain 4,608 images and 6,117 body instances in AnimalPose as the train-set, and 2,000 images and 7,962 body instances in AP-10K as the val-set. Note, we no longer learn to distinguish animal categories.

% Body-hand association for hand tracking on dataset YoutubeHands \cite{huang2022forward}. 
{\bf Downstream Applications:} (1) We select three datasets CrowdHuman val-set \cite{shao2018crowdhuman}, SCUT Head Part\_B \cite{peng2018detecting} and Crowd of Heads Dataset (CroHD) train-set \cite{sundararaman2021tracking} for the {\it accurate crowd head detection} application. The last two datasets do not include body boxes, and were originally released for head detection and dense head tracking tasks, respectively. Thus, it is not feasible to train BPJDet directly on them. Instead, we decide to apply {\it body-head} joint detection models trained on CrowdHuman to them in a cross-domain generalization manner. SCUT Head Part\_B mostly focuses on indoor scenes like classrooms and meeting rooms. CroHD is collected across 9 HD sequences captured from an elevated viewpoint. All sequences are open scenes like crossings and train stations with super high crowd densities. CroHD only provides annotations in 4 train-set sequences. Statistics and samples of them are in Table \ref{dataStat} and Fig. \ref{appCrowdVis}. (2) For the second application {\it hand contact estimation}, we select the dedicated dataset ContactHands \cite{narasimhaswamy2020detecting} for improving performance by leveraging the advanced {\it body-hand} association ability of BPJDet. ContactHands is a large-scale dataset of in-the-wild images for hand detection and contact recognition. It has annotations for 20,506 images, of which 18,877 form the train-set and 1,629 form the test-set. There are 52,050 and 5,983 hand instances in train and test sets, respectively. Some samples are shown in Fig. \ref{appContactVis}.

\setlength{\tabcolsep}{2.1pt} %{0.32pt}
\begin{table}[!t]\scriptsize % \small or \scriptsize or \tiny
	\begin{center}
	\caption{Statistics of first four datasets for body-part joint detection experiments, following two datasets for the accurate crowd head detection task, and the last dataset for the hand contact estimation application. Noting: BodyHands does not distinguish left and right of hand. COCOHumanParts distinguishes left and right of both hand and foot.}
	\label{dataStat}
	\vspace{-5pt}
	\begin{tabular}{l|c|ccccc}
	\Xhline{1.2pt}
	\multirow{2}{*}{Datasets} & \multirow{2}{*}{Images} & \multicolumn{5}{c}{Instance} \\
	\cline{3-7}
	~ & ~ & person & head & face & hand & foot \\
	\Xhline{1.2pt}
	CityPersons \cite{zhang2017citypersons} & 3,475 & 18,201 & 17,954 & 7,922 & --- & --- \\
	CrowdHuman \cite{shao2018crowdhuman} & 19,375 & 439,046 & 439,046 & 248,903 & --- & --- \\
	BodyHands \cite{narasimhaswamy2022whose} & 20,490 & 63,095 & --- & --- & 57,898 & --- \\
	COCOHumanParts \cite{yang2020hier} & 66,808 & 268,030 & 232,392 & 160,102 & 204,827 & 162,099 \\
	%\Xhline{0.8pt}
	\hline
	\hhline{=======} % 7 =
	%\Xhline{0.8pt}
	\hline
	SCUT Head Part\_B \cite{peng2018detecting} & 2,405 & --- & 43,930 & --- & --- & --- \\
	CroHD Train-set \cite{sundararaman2021tracking} & 5,741 & --- & 1,188,793 & --- & --- & --- \\
	\hline
	ContactHands \cite{narasimhaswamy2020detecting} & 20,506 & --- & --- & --- & 58,033 & --- \\
	\Xhline{1.2pt}
	\end{tabular}
	\end{center}
	\vspace{-10pt}
\end{table}

%%%%%%%%%%%%

\subsubsection{Evaluation Metric}
For detection evaluation of body and parts, we report the standard VOC Average Precision (AP) metric with IoU=0.5. We also present the log-average miss rate on False Positive Per Image (FPPI) in the range of [10$^{-2}$, 10$^0$] shortened as MR$^{-2}$ \cite{dollar2011pedestrian} of body and its parts. To assess the association quality of BPJDet, we report the log-average miss matching rate (mMR$^{-2}$) on FPPI of body-part pairs in [10$^{-2}$, 10$^0$], which is originally proposed by BFJDet \cite{wan2021body} for exhibiting the proportion of body-face pairs that are miss-matched:
\begin{equation}\small
	mMR=1 - {N_{mp}}/{N_p} ~
	\label{mmr}
\end{equation}
where $N_{mp}$ is the number of matched body-part pairs with box IoU threshold as 0.5, and $N_p$ represents the total number of pairs. The final value of mMR$^{-2}$ can be naturally acquired by log-averaging all mMR values in the FPPI range. For BPJDet on the body-hand joint detection task, we present Conditional Accuracy and Joint AP defined by BodyHands \cite{narasimhaswamy2022whose} of body-hand pairs. Finally, for body-parts joint detection task on COCOHumanParts, we follow the evaluation protocols defined in Hier R-CNN \cite{yang2020hier}. We report detection performance with a series of APs (AP$_{.5:.95}$, AP$_{.5}$, AP$_{.75}$, AP$_M$, AP$_L$) as in COCO metrics, as well as AP$^{sub}$s based on subordinate relationship which reflect the body-parts association state. Although Hier R-CNN further divides small objects into new small and tiny categories, it does not measure them coherently in the AP$^{sub}$s, we thus abandoned the comparison of AP$_S$ and AP$_S^{sub}$ to show fairness. Evaluations on Animals5C are similar to COCOHumanParts.

For the accurate crowd head detection task, we report various metrics including Precision, Recall, F1 scores and COCO mean Average Precision (mAP) following CroHD \cite{sundararaman2021tracking}. For the hand contact estimation task, following settings in ContactHands \cite{narasimhaswamy2020detecting}, we report four Average Precision (AP) values corresponding to four physical hand-contact states (including No-Contact, Self-Contact, Person-Contact and Object-Contact), and the mAP value of them.

%%%%%%%%%%%%

\subsubsection{Implementation Details}
We adopt the PyTorch 1.10 and 4 RTX-3090 GPUs for training. Depending on the complexity and scale of datasets, we train on the CityPersons, CrowdHuman, BodyHands, COCOHumanParts and Animals5C datasets for 100, 150, 100, 150 and 100 epochs using the SGD optimizer, respectively. Following the original YOLOv5 \cite{yolov5} architecture, we train three kinds of models including BFJDet-S/M/L by controlling the depth and width of bottlenecks in $\mathcal{N}$. All input images are resized and zero-padded to $1536\times1536\times 3$ for being consistent with input shape settings in JointDet \cite{chi2020relational}, BFJDet \cite{wan2021body}, BodyHands \cite{narasimhaswamy2022whose} and Hier R-CNN \cite{yang2020hier}. We adopt effective data training augmentations including mosaic, mix-up and random scaling but leave out test time augmentation (TTA). YOLOv5u \cite{yolov5} and YOLOv8 \cite{yolov8} are similar to the YOLOv5 except the prediction head and loss functions. We only train BPJDet  based on them using the Large parameters BFJDet-L5u and BFJDet-L8u.

As for many hyperparameters, for YOLOv5, we keep most of them unchanged, such as adaptive anchors boxes $\mathcal{B}^s$, the grid balance weight $w_s$, and the loss weights $\alpha\!=\!0.05$, $\beta\!=\!0.7$ and $\gamma\!=\!0.3$. We set $\lambda\!=\!0.015$ based on ablation studies Section \ref{lambdaASs}. For anchor-free detectors YOLOv5u and YOLOv8, the loss weights $\alpha_u\!=\!7.5$, $\beta\!=\!1.5$, $\gamma_u\!=\!0.5$ and $\lambda_u\!=\!1.0/k$. When testing, we use thresholds $\tau^b_{conf}\!=\!0.05$, $\tau^b_{iou}\!=\!0.6$, $\tau^p_{conf}\!=\!0.1$, $\tau^p_{iou}\!=\!0.3$ and $\tau^{inner}_{iou}\!=\!0.6$ for applying $\mathsf{NMS}$ on $\widehat{\mathbf{O}}$. The value of $\tau^{inner}_{iou}$ is a trade-off choice which is discussed in Section \ref{tauAS}. Besides, due to the absence of left and right annotation of hands in BodyHands, we should have to set body-part number $k\!=\!2$ but reduce body-part categories from 2 to 1. For COCOHumanParts and Animals5C, we set body-part number (categories) to 6 and 5, respectively.

%%%%%%%%%%%%%%%%%%%%%%%%

\subsection{Ablation Studies}

\begin{figure*}[t]
	\centering
	\subfloat[]{\includegraphics[width=0.333\textwidth]{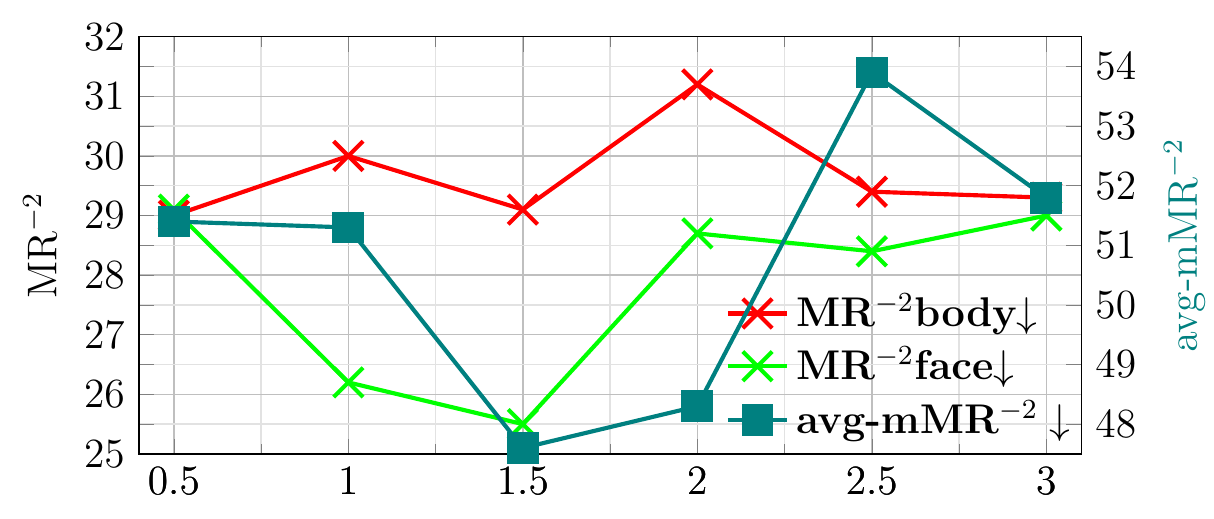}
	\label{lambda}}
	\subfloat[]{\includegraphics[width=0.333\textwidth]{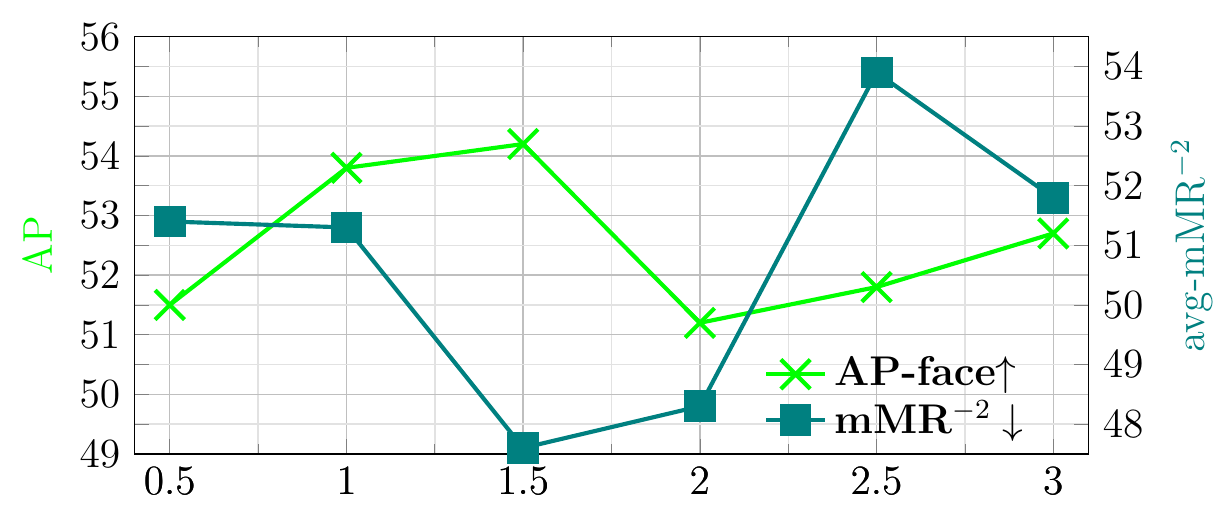}
	\label{lambda1}}
	\subfloat[]{\includegraphics[width=0.333\textwidth]{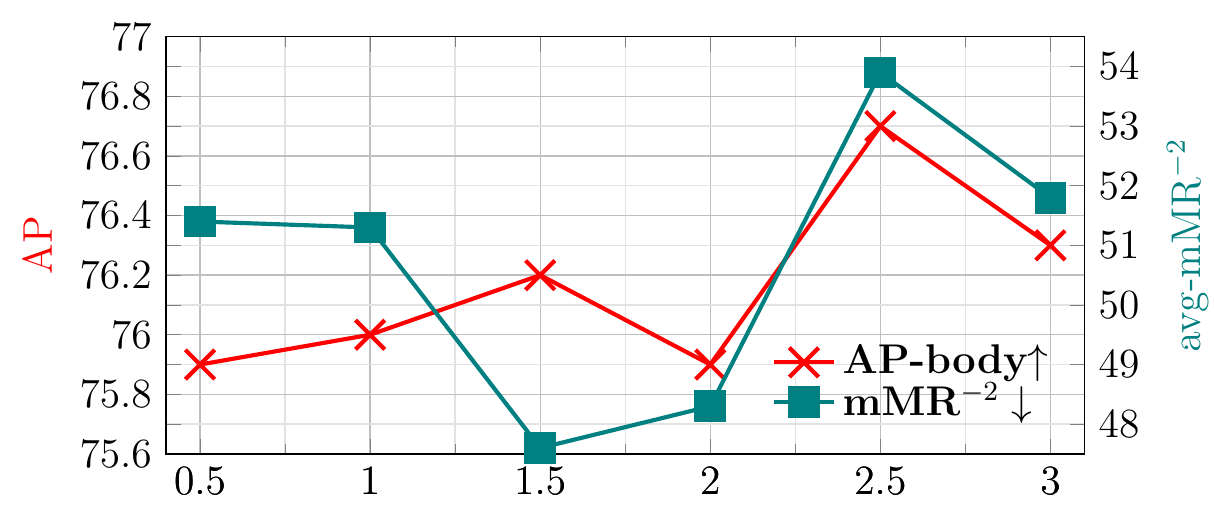}
	\label{lambda2}}
	\vspace{-5pt}
	\caption{The influence of loss weight parameter $\lambda$ (x-axis, enlarged 100$\times$) on pairs of (a) MR$^{-2}$s (body or face) and mMR$^{-2}$, (b) AP-body and mMR$^{-2}$, and (c) AP-face and mMR$^{-2}$.}
	\vspace{-10pt}
	\label{lambdaAS}
\end{figure*}

\begin{figure}[t]
	\centering
	\includegraphics[width=\columnwidth]{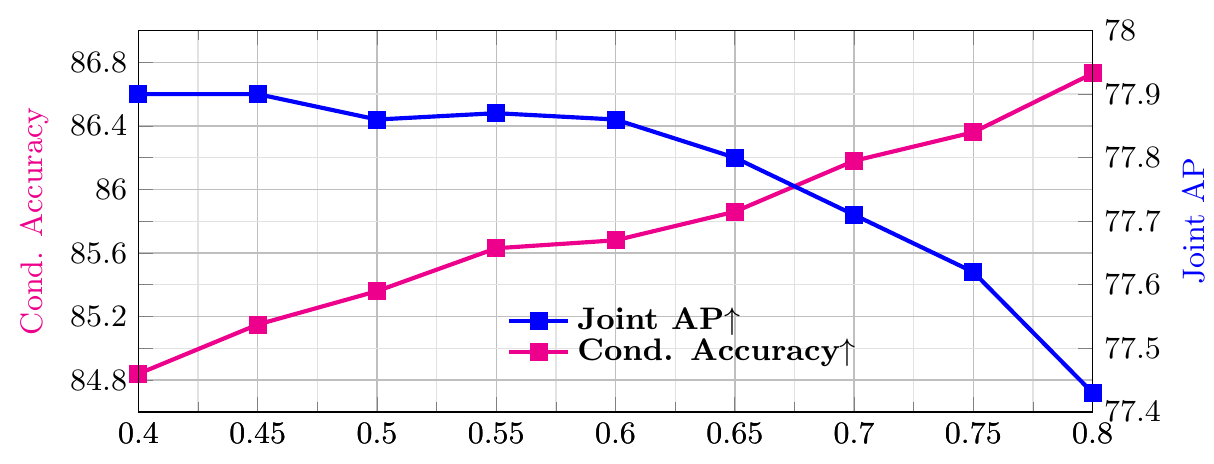}\\
	\vspace{-5pt}
	\caption{The influence of inner IoU threshold $\tau^{inner}_{iou}$ (x-axis).}
	\label{tauInnerIoU}
	\vspace{-10pt}
\end{figure}

\begin{figure}[t]
	\centering
	\includegraphics[width=\columnwidth]{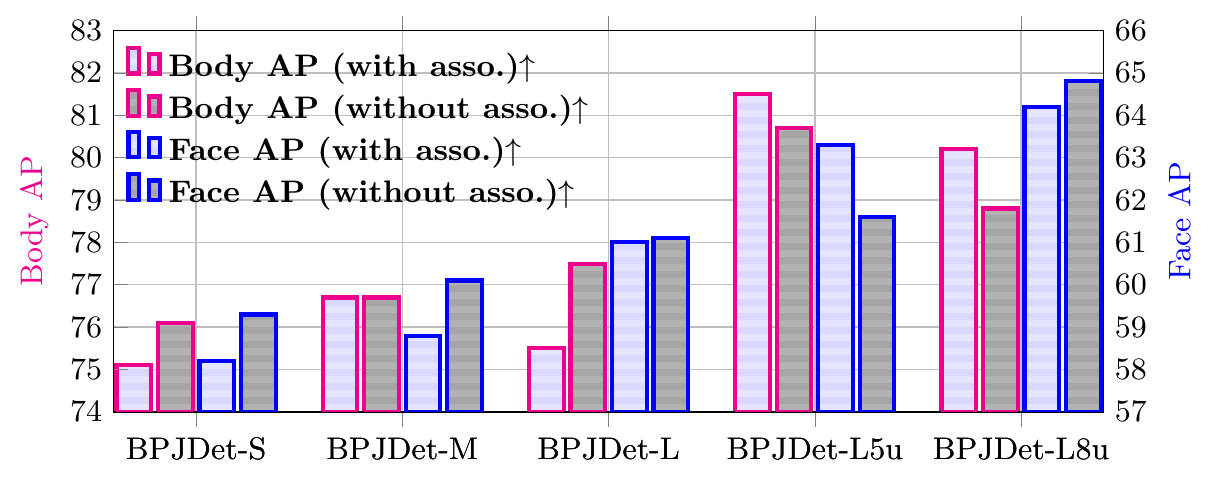}
	\includegraphics[width=\columnwidth]{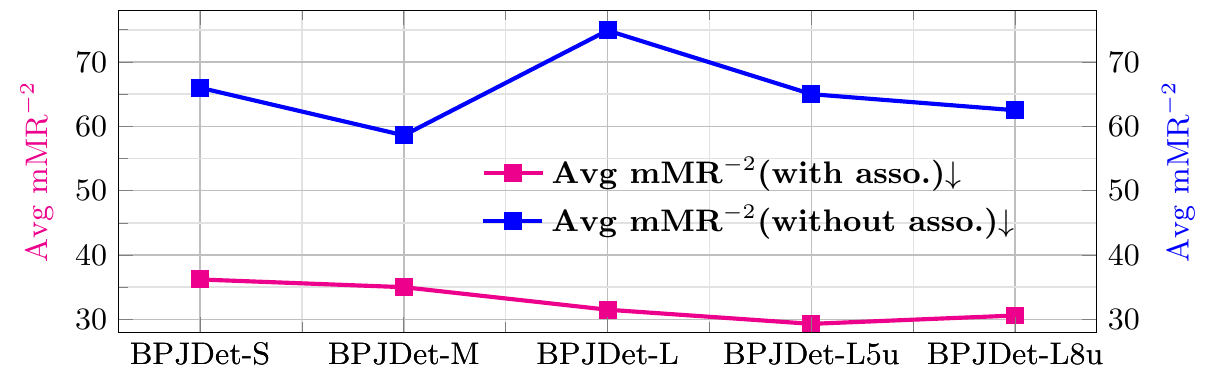}\\
	\vspace{-5pt}
	\caption{The influence of association task on body and part (\eg, face) detection accuracy (\textbf{top figure}) and body-part matching mMR$^{-2}$ (\textbf{bottom figure}) when using different detectors.}
	\label{IoAT}
	\vspace{-10pt}
\end{figure}

\subsubsection{Loss Weight $\lambda$}\label{lambdaASs}
We firstly investigate the hyperparameter $\lambda$. For simplicity, we conduct all ablation experiments on the CityPersons dataset using BPJDet-S and training about joint body-face detection task. The input shape is $1280\times1280\times3$. Total epoch is expanded to 150 for searching the best result. We uniformly sample $\lambda$ from 0.005 to 0.030 with step 0.005 for training, and report metrics including MR$^{-2}$s and average mMR$^{-2}$ of each best model. As in Fig.~\ref{lambda}, we can clearly find that the model performance is optimal when $\lambda\!=\!0.015$. A larger or smaller $\lambda$ will lead to inferior results. 

In addition to investigating the influence of $\lambda$ on MR$^{-2}$ and average mMR$^{-2}$, we also report the metric AP for body and face to reflect its impact. As shown in Fig.~\ref{lambda1}, we still obtain the optimal AP for face when $\lambda\!=\!0.015$. Although we do not get the best AP for body as shown in Fig.~\ref{lambda2} by setting $\lambda\!=\!0.015$, we decide to give priority to ensuring the accuracy of the body-part matching after the trade-off between detection and association.

\subsubsection{Inner IoU Threshold $\tau^{inner}_{iou}$}\label{tauAS}
In the Section \ref{inference}, we introduce a new and vital threshold $\tau^{inner}_{iou}$ for better allocating part boxes to the body boxes they belong to. We conduct the threshold searching task on the BodyHands~\cite{narasimhaswamy2022whose} val-set using the already trained body-hand joint detection model BPJDet-S. As shown in Fig.~\ref{tauInnerIoU}, we select $\tau^{inner}_{iou}$ from 0.4 to 0.8 with a step 0.05, and observe the change trend of indicators Cond. Accuracy and Joint AP. Finally, we set $\tau^{inner}_{iou}\!=\!0.6$ for a better trade-off between conditional accuracy and joint AP.

\subsubsection{Influence of Association Task}\label{IoATAS}
To investigate the influence of extended representation for body-part association task on detection accuracy, we additionally trained all models including BPJDet-S/M/L/L5u/L8u on the CityPersons dataset while removing the association task ($\lambda\!=\!\lambda_u\!=\!0$). All AP results of body and part (face) are shown in Fig.~\ref{IoAT} top. We can observe that regardless of the type of detector used, there is no significant difference in the detection accuracy before and after the introduction of extended representation, \ie, the addition of body-part association in BPJDet. Even in some cases, especially for anchor-free detectors, the association task may lead to higher detection performance. These results indicate that our extended proposed representation is indeed harmless and not sensitive about backbones. Besides, we also evaluated the average mMR$^{-2}$ of BPJDet when not using our extended representation for jointly learning the association task. As shown in Fig.~\ref{IoAT} bottom, the trivial body-part matching results of them are very poor, which reveals the necessity of association design.

%%%%%%%%%%%%%%%%%%%%%%%%
\subsection{Quantitative and Visual Comparison}\label{QVC}

\subsubsection{Body-Face Joint Detection}
We compare joint body-face detection performance of our BPJDet with method BFJDet \cite{wan2021body} on two benchmarks: CityPersons and CrowdHuman.

{\bf CityPersons:} The Table \ref{CityPersons} shows results on the val-set of CityPersons. Following strategies in \cite{wang2018repulsion, wan2021body}, results are reported on four subsets of {\it Reasonable} (occlusion $<$ 35\%), {\it Partial} (10\% $<$ occlusion $\leqslant$ 35\%), {\it Bare} (occlusion $\leqslant$ 10\%) and {\it Heavy} (occlusion $>$ 35\%). We also report the average value ({\it Average}) of these four results. For BPJDet using YOLOv5, comparing with RetinaNet+BFJ which obtains similar body AP but far inferior face AP to ours, BPJDet-L exceeds it of mMR$^{-2}$ by {\bf 13.1}\%, {\bf 13.8}\%, {\bf 13.0}\% and {\bf 16.9}\% in four subsets, respectively. Comparing with FPN+BFJ that has similar face AP but higher body AP to ours, BPJDet-L also surpasses it of mMR$^{-2}$ by {\bf 6.3}\%, {\bf 2.9}\%, {\bf 7.5}\% and {\bf 7.3}\% in four subsets. Especially, in the {\it Heavy} subset, our association superiority is the most prominent, which reveals that our BPJDet has an advantage for addressing miss-matchings in crowded scenes. When using advanced anchor-free detectors which achieved higher AP results of body and face than BPJDet-L, BPJDet-L5u and BPJDet-L8u can further improve the association ability in four subsets. We also obtained two new lower {\it Average} mMR$^{-2}$ results {\bf 29.3}\% and {\bf 30.6}\% than 31.5\% of the anchor-based BPJDet-L. This proves the generalization of extended representation.

\setlength{\tabcolsep}{2.1pt} %{0.32pt}
\begin{table}[!t]\scriptsize % \small or \scriptsize or \tiny
	\begin{center}
	%\caption{The performance comparison of the joint {\it body-face} detection task in the val-set of CityPersons dataset.}
	\caption{Results of the joint {\it body-face} detection in the val-set of CityPersons.}
	\label{CityPersons}
	\vspace{-5pt}
	\begin{tabular}{l|c|c|cccc|c}
	\Xhline{1.2pt}
	\multirow{2}{*}{Methods} & \multirow{2}{*}{\makecell{AP$\uparrow$\\body}} & \multirow{2}{*}{\makecell{AP$\uparrow$\\face}} &  \multicolumn{5}{c}{body-face mMR$^{-2}\downarrow$} \\
	\cline{4-8}
	~ & ~ & ~ & {\it Reasonable} & {\it Partial} & {\it Bare} & {\it Heavy} & {\it Average} \\
	\Xhline{1.2pt}
	RetinaNet+POS \cite{lin2017focal, wan2021body} & 78.5 & 35.3 & 40.0 & 42.8 & 38.7 & 67.0 & 47.1 \\
	RetinaNet+BFJ \cite{lin2017focal, wan2021body} & 79.3 & 36.2 & 39.5 & 41.5 & 38.5 & 63.1 & 45.7 \\
	FPN+POS \cite{lin2017feature, wan2021body} & 80.6 & 65.5 & 33.5 & 32.7& 34.1 & 56.6 & 39.2 \\
	FPN+BFJ \cite{lin2017feature, wan2021body} & {\bf 84.4} & {\bf 68.0} & 32.7 & 30.6 & 33.0 & 53.5 & 37.5 \\
	\hline
	BPJDet-S (Ours) & 75.1 & 58.2 & 29.3 & 28.9 & 29.3 & 57.2 & 36.2 \\
	BPJDet-M (Ours) & 76.7 & 58.8 & 27.5 & 31.6 & 24.9 & 55.8 & 35.0 \\
	BPJDet-L (Ours) & 75.5 & 61.0 & 26.4 & 27.7 & 25.5 & {\bf 46.2} & 31.5 \\
	BPJDet-L5u (Ours) & 81.5 & 63.3 & {\bf 22.9} & 24.0 & {\bf 22.1} & 48.4 & {\bf 29.3}  \\
	BPJDet-L8u (Ours) & 80.2 & 64.2 & 24.3 & {\bf 21.1} & 26.1 & 50.9 & 30.6 \\
	\Xhline{1.2pt}
	\end{tabular}
	\end{center}
	\vspace{-10pt}
\end{table}

{\bf CrowdHuman:} The Table \ref{CrowdHuman} shows results on the more challenging CrowdHuman. Our method BPJDet using either anchor-based YOLOv5 or anchor-free detectors (YOLOv5u and YOLOv8) achieves considerable gains in all metrics comparing with BFJDet based on one-stage RetinaNet \cite{lin2017focal} or two-stage FPN \cite{lin2017feature} and CrowdDet \cite{chu2020detection}. Remarkably, BPJDet-L8u has achieved the lowest mMR$^{-2}$ value {\bf 49.9}\%, which is {\bf 2.4}\% lower than the previous best method CrowdDet+BFJ. Moreover, our BPJDet surpasses CrowdDet+BFJ on body-face association with keeping higher detection performance of body and face, which means dealing with more instance matching issues. These again demonstrate the universality and superiority of our method. % More persuasive visual results of BPJDet-L trained on CrowdHuman and tested on complicated overcrowded scenes are shown in Fig. \ref{moreExamples} %Fig. \ref{BPJDetFace}.

\setlength{\tabcolsep}{3.2pt} %{0.7pt}
\begin{table}[!t]\scriptsize % \small or \scriptsize or \tiny
	\begin{center}
	%\caption{The performance comparison of the joint {\it body-face} detection task in the val-set of CrowdHuman dataset.}
	\caption{Results of the joint {\it body-face} detection in the val-set of CrowdHuman.}
	\label{CrowdHuman}
	\vspace{-5pt}
	\begin{tabular}{l|c|c|c|c|c|c}
	\Xhline{1.2pt}
	\multirow{2}{*}{Methods} & \multirow{2}{*}{Stage} & \multirow{2}{*}{\makecell{MR$^{-2}$\\body$\downarrow$}} & \multirow{2}{*}{\makecell{MR$^{-2}$\\face$\downarrow$}} & \multirow{2}{*}{\makecell{AP$\uparrow$\\body}} & \multirow{2}{*}{\makecell{AP$\uparrow$\\face}} & \multirow{2}{*}{\makecell{body-face\\mMR$^{-2}\downarrow$}} \\
	~ & ~ & ~ & ~ & ~ & ~ \\
	\Xhline{1.2pt}
	RetinaNet+POS \cite{lin2017focal, wan2021body} & One & 52.3 & 60.1 & 79.6 & 58.0 & 73.7 \\
	RetinaNet+BFJ \cite{lin2017focal, wan2021body} & One & 52.7 & 59.7 & 80.0 & 58.7 & 63.7 \\ 
	FPN+POS \cite{lin2017feature, wan2021body} & Two & 43.5 & 54.3 & 87.8 & 70.3 & 66.0 \\
	FPN+BFJ \cite{lin2017feature, wan2021body} & Two & 43.4 & 53.2 & 88.8 & 70.0 & 52.5 \\
	CrowdDet+POS \cite{chu2020detection, wan2021body} & Two & 41.9 & 54.1 & 90.7 & 69.6 & 64.5 \\ 
	CrowdDet+BFJ \cite{chu2020detection, wan2021body} & Two & 41.9 & 53.1 & 90.3 & 70.5 & 52.3 \\ 
	\hline
	BPJDet-S (Ours) & One & 41.3 & 45.9 & 89.5 & 80.8 & 51.4 \\
	BPJDet-M (Ours) & One & 39.7 & {\bf 45.0} & 90.7 & {\bf 82.2} & 50.6 \\
	BPJDet-L (Ours) & One & 40.7 & 46.3 & 89.5 & 81.6 & 50.1 \\
	BPJDet-L5u (Ours) & One & 37.0 & 45.1 & {\bf 91.8} & 81.9 & 50.4 \\
	BPJDet-L8u (Ours) & One & {\bf 36.9} & {\bf 45.0} & 91.5 & 82.1 & {\bf 49.9} \\
	\Xhline{1.2pt}
	\end{tabular}
	\end{center}
	\vspace{-10pt}
\end{table}

%%%%%%%%%%%

\subsubsection{Body-Hand Joint Detection}
\begin{figure}[]
	\includegraphics[height = 0.280\columnwidth]{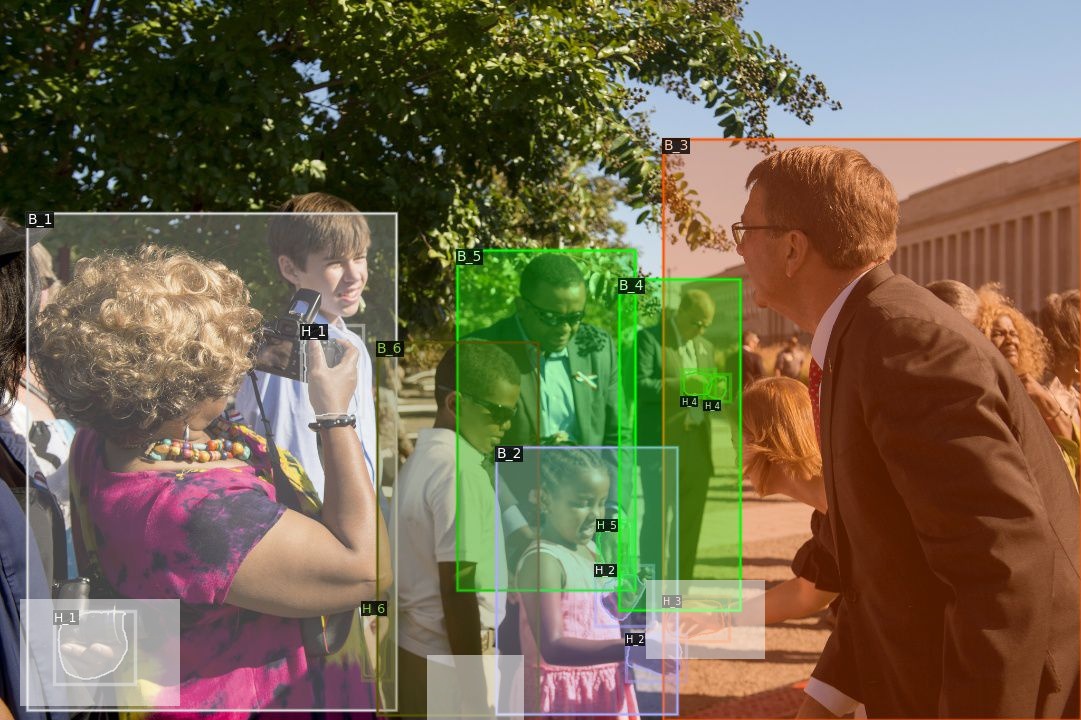}
	\includegraphics[height = 0.280\columnwidth]{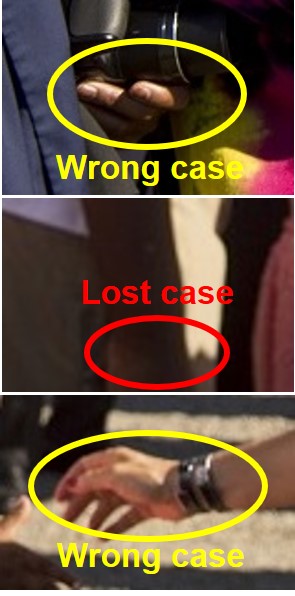}
 	\includegraphics[height = 0.280\columnwidth]{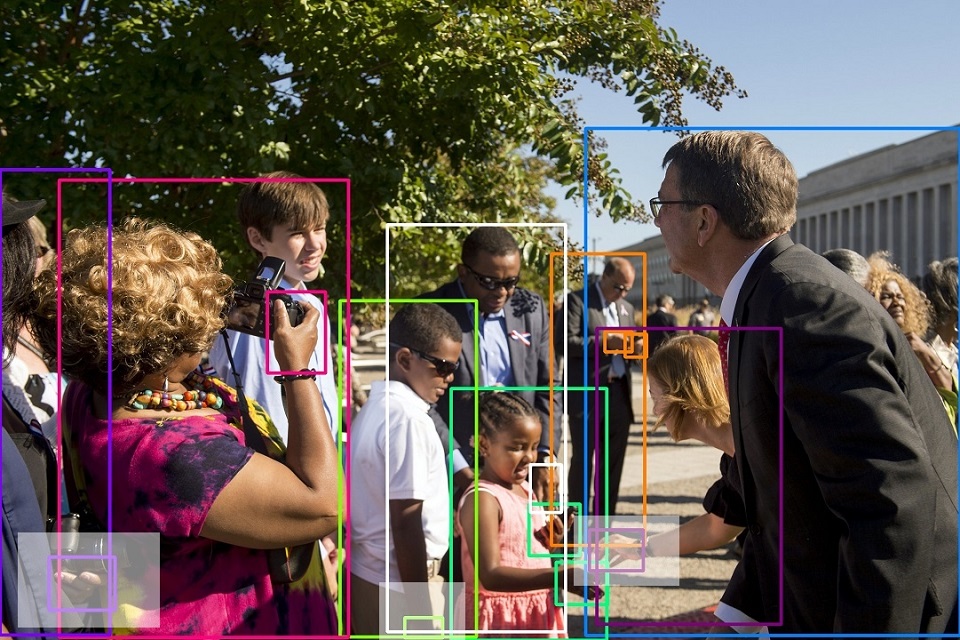}\\
	\vspace{-10pt}\\
	\includegraphics[height = 0.262\columnwidth]{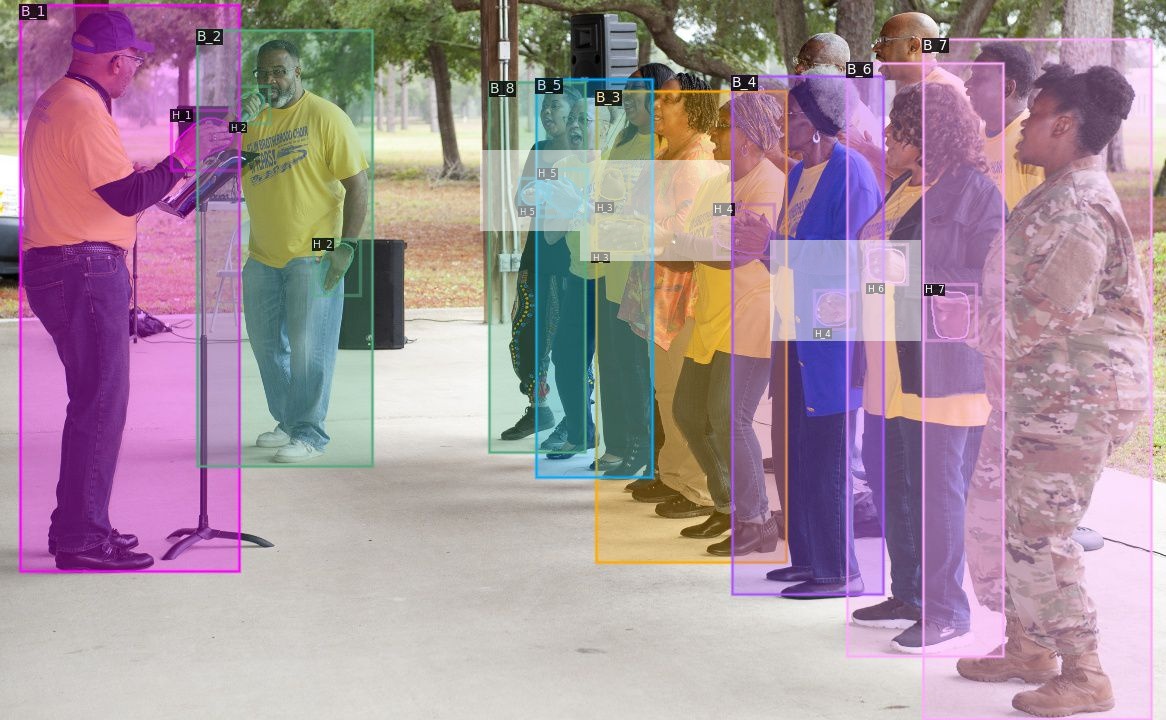}
	\includegraphics[height = 0.262\columnwidth]{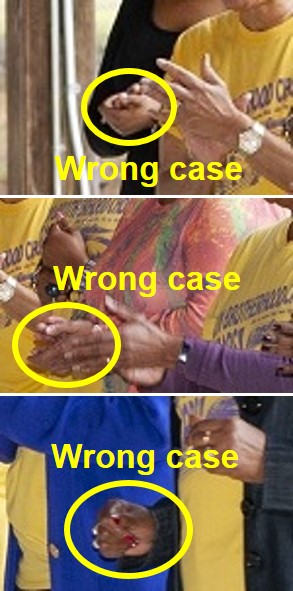}
 	\includegraphics[height = 0.262\columnwidth]{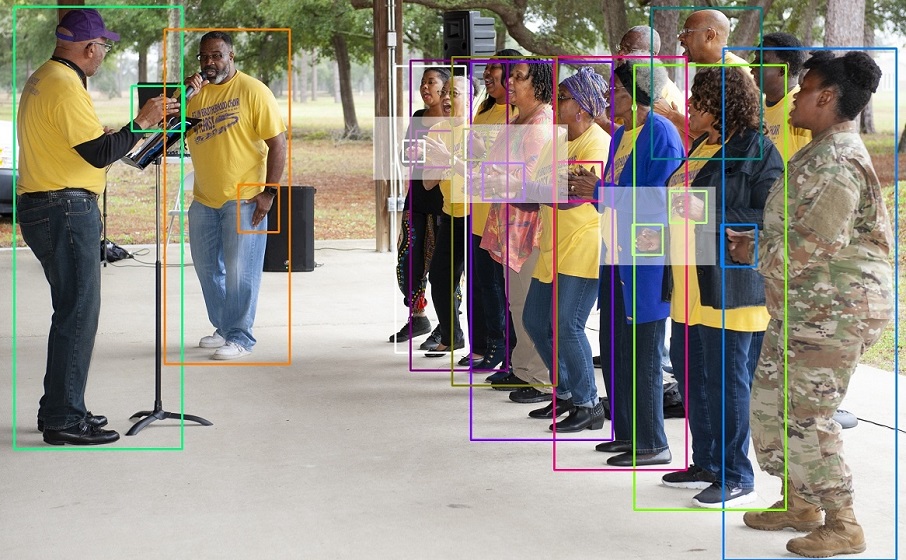}\\
	\vspace{-10pt}\\
	\includegraphics[height = 0.278\columnwidth]{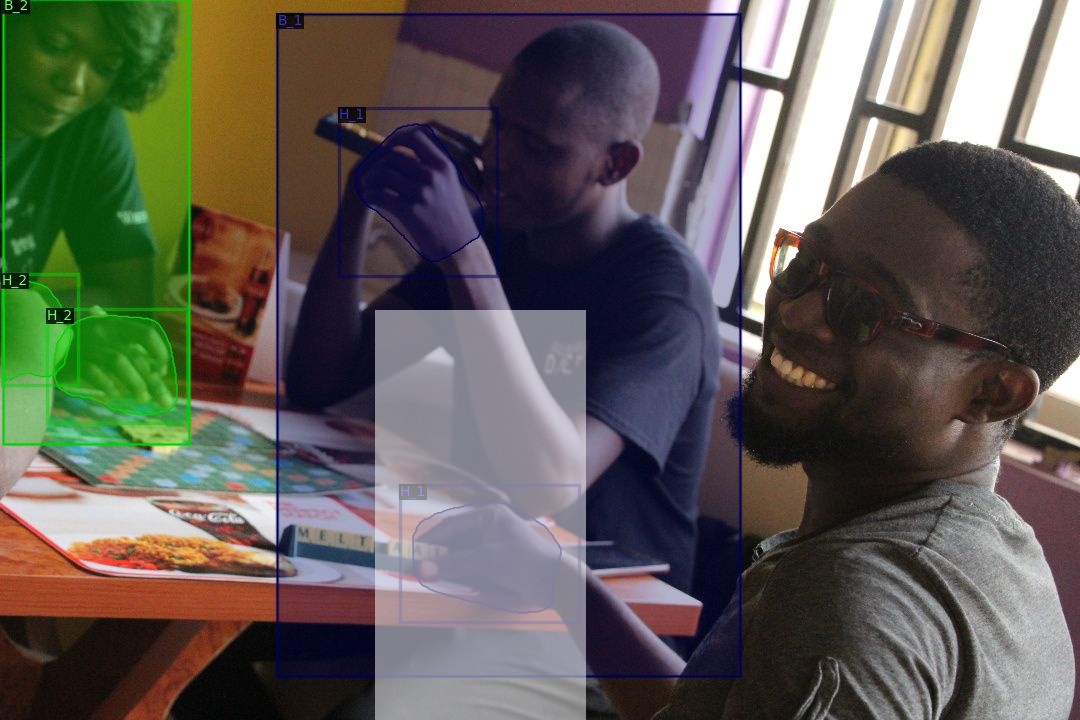}
	\includegraphics[height = 0.278\columnwidth]{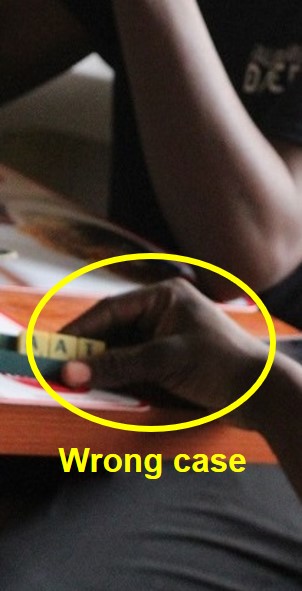}
	\includegraphics[height = 0.278\columnwidth]{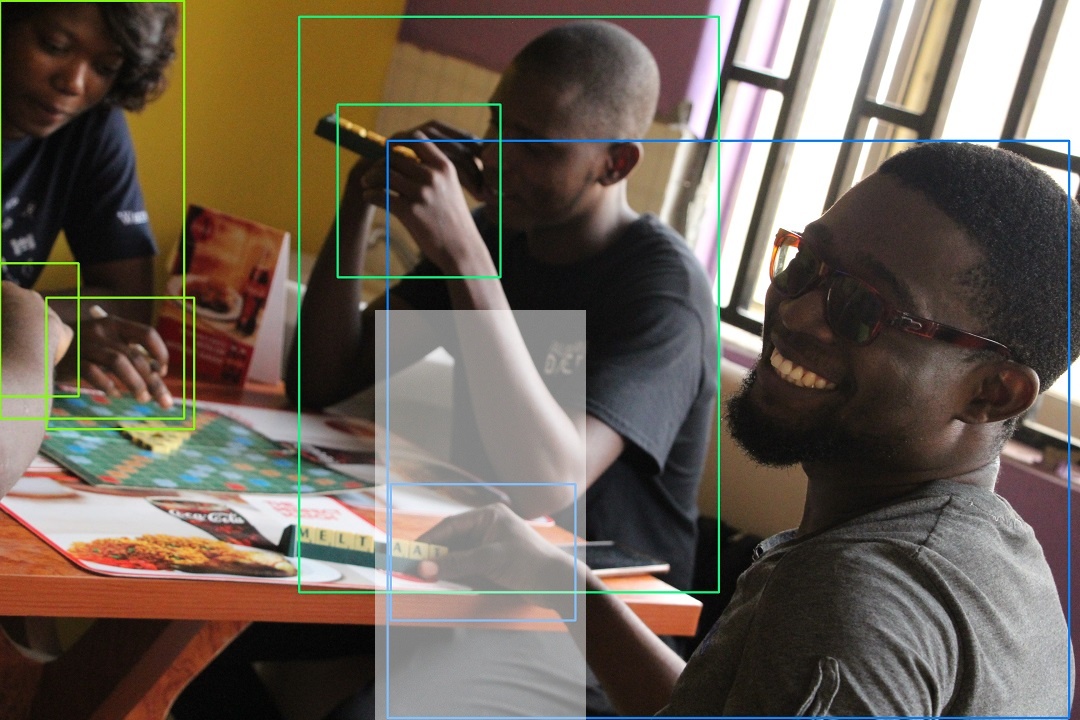}
	\caption{Qualitative results comparison of our BPJDet-L ({\bf right column}) for the joint {\it body-hand} detection task with the method in BodyHands ({\bf left column}) on its val-set images. These failure cases ({\bf middle column}) of method in BodyHands are directly fetched from its paper. We masked them with white transparent boxes for quick inspection.}
	\label{BPJDetHand}
	\vspace{-10pt}
\end{figure}

We conduct joint body-hand detection experiments on BodyHands \cite{narasimhaswamy2022whose}, and compare our BPJDet with several methods proposed in BodyHands. As shown in Table \ref{BodyHands}, our BPJDet using either anchor-based or anchor-free detectors outperforms all other methods by a significant margin. With achieving highest hand AP 87.6\% and conditional accuracy 86.93\% of body, BPJDet-L5u largely improves the previous best Joint AP of body and hands by {\bf 21.55}\%. We give more qualitative comparisons of joint body-hand detection trained on BodyHands in Fig. \ref{BPJDetHand}. The compared masked images are fetched from BodyHands paper. Our BPJDet can detect many unlabeled objects, and associate hands that method proposed in BodyHands failed to match. This illustrates why we have an overwhelming Joint AP advantage. All these results further validate the generalizability of our extended object representation and its impressive strength on body and part relationship discovery. 

\setlength{\tabcolsep}{2.7pt} %{0.4pt}
\begin{table}[!t]\scriptsize  % \small or \scriptsize or \tiny
	\begin{center}
	%\caption{The performance comparison of the joint {\it body-hand} detection task in the val-set of BodyHands. The FD, FS and LD means Feature Distance, Feature Similarity and Location Distance, respectively. The marker * means using hand self-association option.}
	\caption{Results of the joint {\it body-hand} detection in the val-set of BodyHands. The FD, FS and LD means Feature Distance, Feature Similarity and Location Distance, respectively. The * means hand self-association.}
	\label{BodyHands}
	\vspace{-5pt}
	\begin{tabular}{l|c|c|c}
	\Xhline{1.2pt}
	Methods & Hand AP$\uparrow$ & Cond. Accuracy$\uparrow$ & Joint AP$\uparrow$ \\
	\Xhline{1.2pt}
	OpenPose \cite{cao2017realtime} & 39.7 & 74.03 & 27.81 \\
	Keypoint Communities \cite{zauss2021keypoint} & 33.6 & 71.48 & 20.71 \\
	MaskRCNN+FD \cite{he2017mask, narasimhaswamy2022whose} & 84.8 & 41.38 & 23.16 \\
	MaskRCNN+FS \cite{he2017mask, narasimhaswamy2022whose} & 84.8 & 39.12 & 23.30 \\
	MaskRCNN+LD \cite{he2017mask, narasimhaswamy2022whose} & 84.8 & 72.83 & 50.42 \\
	MaskRCNN+IoU \cite{he2017mask, narasimhaswamy2022whose} & 84.8 & 74.52 & 51.74 \\
	BodyHands \cite{narasimhaswamy2022whose} & 84.8 & 83.44 & 63.48 \\
	BodyHands* \cite{narasimhaswamy2022whose} & 84.8 & 84.12 & 63.87 \\
	\hline
	BPJDet-S (Ours) & 84.0 & 85.68 & 77.86 \\
	BPJDet-M (Ours) & 85.3 & 86.80 & 78.13 \\
	BPJDet-L (Ours) & 85.9 & 86.91 & 84.39 \\
	BPJDet-L5u (Ours) & {\bf 87.6} & {\bf 86.93} & {\bf 85.42} \\
	BPJDet-L8u (Ours) & 86.8 & 85.07 & 78.44 \\
	\Xhline{1.2pt}
	\end{tabular}
	\end{center}
	\vspace{-10pt}
\end{table}

%%%%%%%%%%%

\subsubsection{Body-Head Joint Detection}\label{subsecBHJD}
To further verify the strong adaptability of our proposed BPJDet, we implement the joint body-head detection test on the CrowdHuman~\cite{shao2018crowdhuman} dataset, and compare our body AP and MR$^{-2}$ results with popular pedestrian detection methods towards crowded scenes. All experiment settings are the same as training of body-face pairs. The results are shown in Tabel~\ref{CrowdHumanHead}. In order to compare its association difference with the joint body-face detection task, we replicate the body face pair detection results of BPJDet for reference.

We can observe that, comparing with the body-face joint detection task, both of the AP and MR$^{-2}$ results become worse, although the mMR$^{-2}$ shows a better body-head association effect (improving from 49.9 to 46.1). On one hand, this may reflect that head detection is harder than face detection due to showing less appearance feature when face area is invisible. On the other hand, training body detection jointly with a challenging body-part (e.g., full-view head) may be a burden on its AP (dropping from the best result 91.8 to 87.4) and MR$^{-2}$ (dropping from the best result 36.9 to 41.1) performance. Nevertheless, compared with the previous best body-head joint detection methods, including DA-RCNN \cite{zhang2019double} and JointDet \cite{chi2020relational}, we still present considerable advantages in MR$^{-2}$ performance. Finally, we also reported SOTA methods focusing on independent detection of body objects in crowded people. The so far lowest MR$^{-2}$ of body is maintained by PedHunter \cite{chi2020pedhunter}. Our BPJDet-L8u model for body-face joint detection task achieves a lower result with it (36.9 vs. 39.5). The best AP of body is held by Iter-E2EDET \cite{zheng2022progressive}, which is based on the cumbersome DETR. However, such query-based methods are slightly inferior to their box-based counterparts in performance of body MR$^{-2}$. We consider alleviating this drawback by exploring the DETR for body-part joint detection in future.

\setlength{\tabcolsep}{2.1pt} %{0.7pt}
\begin{table}[!t]\scriptsize % \small or \scriptsize or \tiny
	\begin{center}
	\caption{The performance comparison of our method about the joint {\it body-head} detection task with other crowd person detection methods in the val-set of CrowdHuman. The marker $\dagger$ means recent query-based detection methods rather than other box-based or point-based methods.}
	\label{CrowdHumanHead}
	\vspace{-5pt}
	\begin{tabular}{l|c|c|c|c|c|c}
	\Xhline{1.2pt}
	\multirow{2}{*}{Methods} & \multirow{2}{*}{\makecell{Joint\\Detect?}} & \multirow{2}{*}{\makecell{MR$^{-2}$\\body$\downarrow$}} & \multirow{2}{*}{\makecell{MR$^{-2}$\\{\bf head}$\downarrow$}} & \multirow{2}{*}{\makecell{AP$\uparrow$\\body}} & \multirow{2}{*}{\makecell{AP$\uparrow$\\{\bf head}}} & \multirow{2}{*}{\makecell{body-head\\mMR$^{-2}\downarrow$}} \\
	~ & ~ & ~ & ~ & ~ & ~ \\
	\Xhline{1.2pt}
	CrowdHuman baseline \cite{shao2018crowdhuman} & \xmark & 50.4 & 52.1 & 85.0 & 78.0 & --- \\  % FPN based
	Adaptive-NMS \cite{liu2019adaptive} & \xmark & 49.7 & --- & 84.7 & --- & --- \\
	PBM \cite{huang2020nms} & \xmark & 43.3 & --- & 89.3 & --- & --- \\
	CrowdDet \cite{chu2020detection} & \xmark & 41.4 & --- & 90.7 & --- & --- \\
	AEVB \cite{zhang2021variational} & \xmark & 40.7 & --- & --- & --- & --- \\ % Faster R-CNN based
	AutoPedestrian \cite{tang2021autopedestrian} & \xmark & 40.6 & --- & --- & --- & --- \\  % CrowdDet based
	Beta RCNN(KL$_{th}$ = 7) \cite{xu2020beta} & \xmark & 40.3 & --- & 88.2 & --- & --- \\
	PedHunter \cite{chi2020pedhunter} & \xmark & {\bf 39.5} & --- & --- & --- & --- \\
	\hline
	Sparse-RCNN \cite{sun2021sparse}$\dagger$ & \xmark & 44.8 & --- & 91.3 & --- & --- \\
	Deformable-DETR \cite{zhu2020deformable}$\dagger$ & \xmark & 43.7 & --- & 91.5 & --- & --- \\
	PED-DETR \cite{lin2020detr}$\dagger$ & \xmark & 43.7 & --- & 91.6 & --- & --- \\
	Iter-E2EDET \cite{zheng2022progressive}$\dagger$ & \xmark & 41.6 & --- & {\bf 92.5} & --- & --- \\
	\Xhline{0.8pt}
	DA-RCNN \cite{zhang2019double} & \cmark & 52.3 & 50.0 & --- & --- & --- \\
	DA-RCNN + J-NMS \cite{zhang2019double} & \cmark & 51.8 & 49.7 & --- & --- & --- \\
	JointDet w/o RDM \cite{chi2020relational} & \cmark & 47.0 & 48.7 & --- & --- & --- \\
	JointDet \cite{chi2020relational} & \cmark & 46.5 & 48.3 & --- & --- & --- \\
	\hline
	BPJDet-S (Ours) & \cmark & 45.9 & 47.8 & 85.0 & 79.4 & 48.0 \\
	BPJDet-M (Ours) & \cmark & 45.4 & {\bf 46.8} & 85.3 & {\bf 80.5} & {\bf 46.1} \\
	BPJDet-L (Ours) & \cmark & 46.2 & 48.0 & 83.8 & 78.6 & 46.4 \\
	BPJDet-L5u (Ours) & \cmark & 41.1 & 50.2 & 87.4 & 79.9 & 49.9 \\
	BPJDet-L8u (Ours) & \cmark & 41.8 & 50.4 & 86.6 & 79.8 & 49.4 \\
	\Xhline{0.8pt}
	\hhline{=======} % 7 =
	\Xhline{0.8pt}
	\multirow{2}{*}{Methods} & \multirow{2}{*}{\makecell{Joint\\Detect?}} & \multirow{2}{*}{\makecell{MR$^{-2}$\\body$\downarrow$}} & \multirow{2}{*}{\makecell{MR$^{-2}$\\{\bf face}$\downarrow$}} & \multirow{2}{*}{\makecell{AP$\uparrow$\\body}} & \multirow{2}{*}{\makecell{AP$\uparrow$\\{\bf face}}} & \multirow{2}{*}{\makecell{body-face\\mMR$^{-2}\downarrow$}} \\
	~ & ~ & ~ & ~ & ~ & ~ \\
	\Xhline{1.2pt}
	BPJDet-S (Ours) & \cmark & 41.3 & 45.9 & 89.5 & 80.8 & 51.4 \\
	BPJDet-M (Ours) & \cmark & 39.7 & {\bf 45.0} & 90.7 & {\bf 82.2} & 50.6 \\
	BPJDet-L (Ours) & \cmark & 40.7 & 46.3 & 89.5 & 81.6 & 50.1 \\
	BPJDet-L5u (Ours) & \cmark & 37.0 & 45.1 & {\bf 91.8} & 81.9 & 50.4 \\
	BPJDet-L8u (Ours) & \cmark & {\bf 36.9} & {\bf 45.0} & 91.5 & 82.1 & {\bf 49.9} \\
	\Xhline{1.2pt}
	\end{tabular}
	\end{center}
	\vspace{-10pt}
\end{table}

%%%%%%%%%%%

\subsubsection{Body-Parts Joint Detection}\label{subsecBPJD}
We conduct joint body-parts detection experiments on the COCOHumanParts \cite{yang2020hier} dataset to explain that our generic BPJDet is not limited to body part numbers and categories. All results are shown in Table \ref{HumanParts}. On the premise of similar parameters, our YOLOv5-S based model BPJDet-S is far better than a series of ResNet-50 based baselines (e.g., Faster R-CNN \cite{ren2015faster}, Mask R-CNN \cite{he2017mask} RetinaNet \cite{lin2017focal}, FCOS \cite{tian2019fcos} and Hier R-CNN \cite{yang2020hier}) in terms of all and per categories of AP metrics. Although Mask R-CNN \cite{he2017mask} and Hier R-CNN \cite{yang2020hier} employ backbones with larger parameters like ResNet-101 and ResNeXt-101, our BPJDet-S still maintains a leading advantage over them. After incorporating various enhanced components, including high-capacity backbones, deformable convs \cite{dai2017deformable}, and training time augmentation, Hier R-CNN achieves its best results of all categories APs (Hier-X101$\dagger\ddag$) that remain comparable to BPJDet-M/L/L5u/L8u narrowly. And note that the cost is more complex structural design and greater computation. In terms of detection accuracy, we reveal the necessity and urgency of exploring advanced detectors like YOLOv5 and YOLOv8, rather than the ancient Faster R-CNN and its variants.

\begin{figure}[]
	\includegraphics[width=0.495\columnwidth]{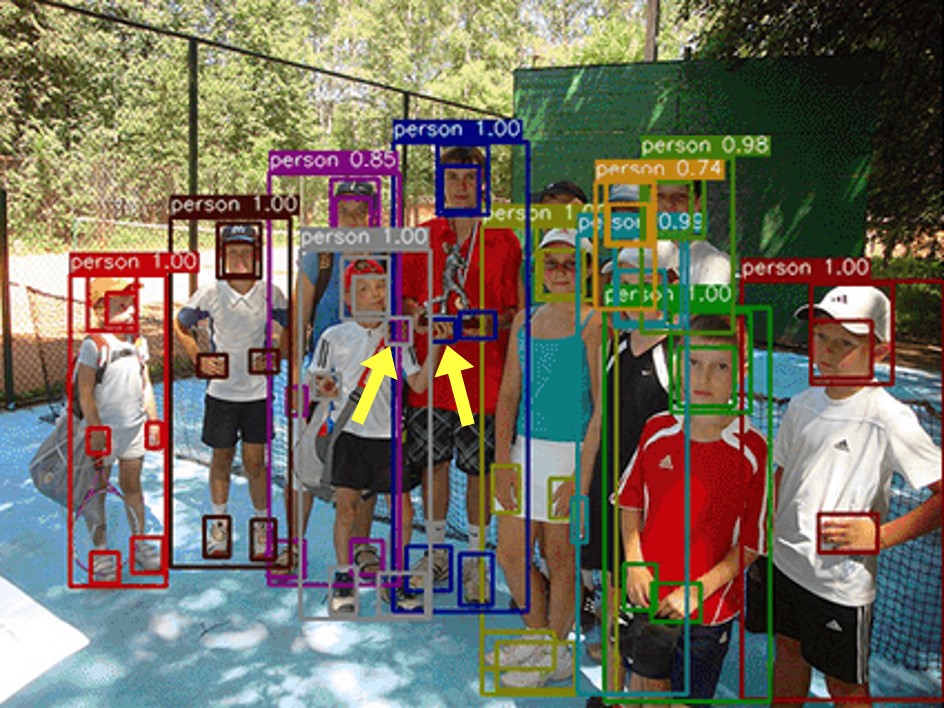}
	\includegraphics[width=0.495\columnwidth]{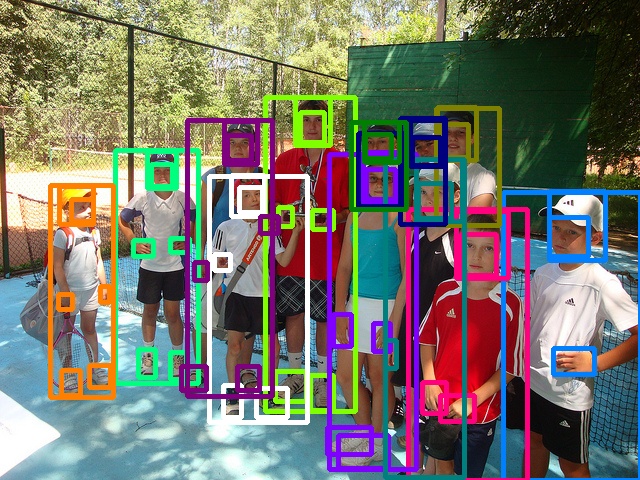}\\
	\vspace{-9pt}\\
	\includegraphics[width=0.495\columnwidth]{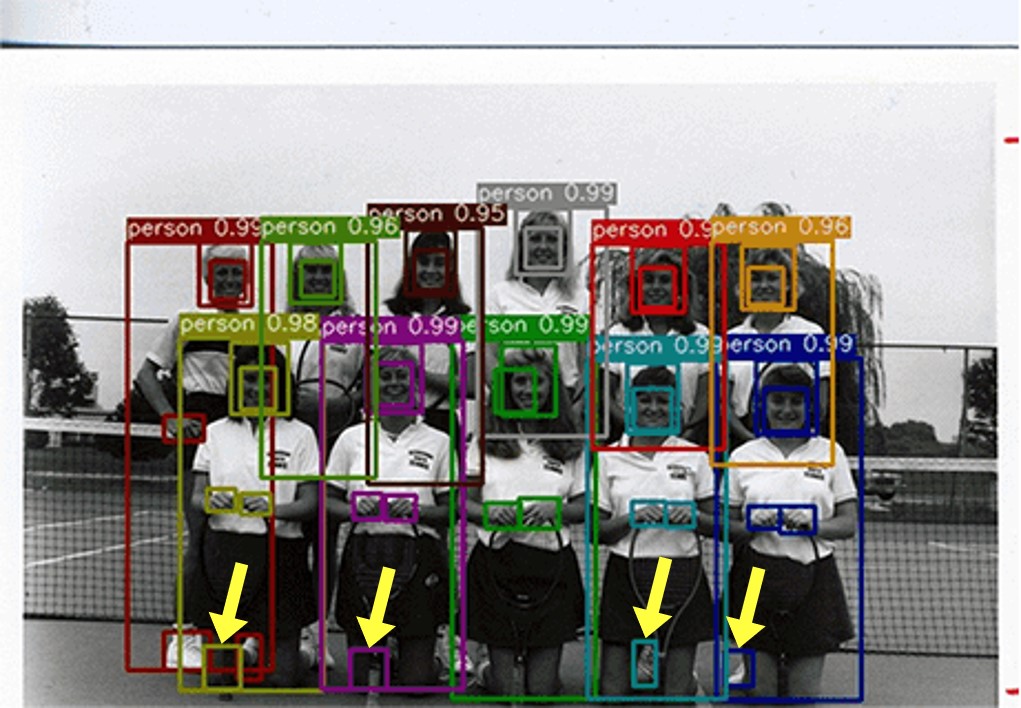}
 	\includegraphics[width=0.495\columnwidth]{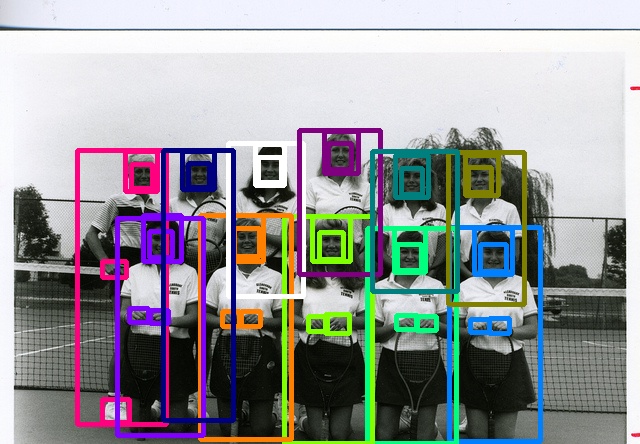}\\
	\vspace{-9pt}\\
	\includegraphics[width=0.495\columnwidth]{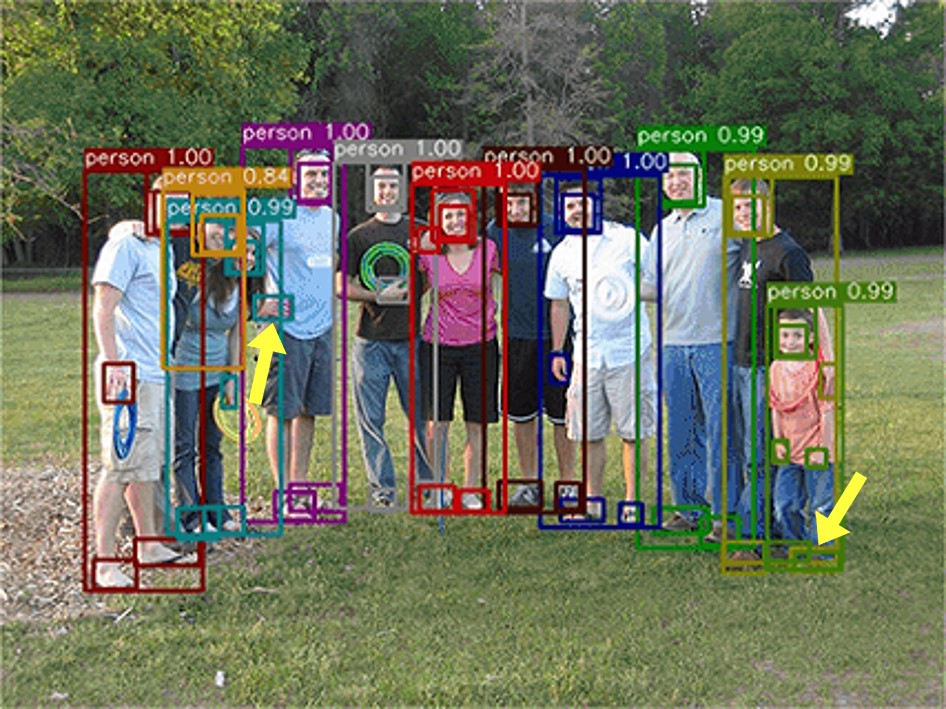}
 	\includegraphics[width=0.495\columnwidth]{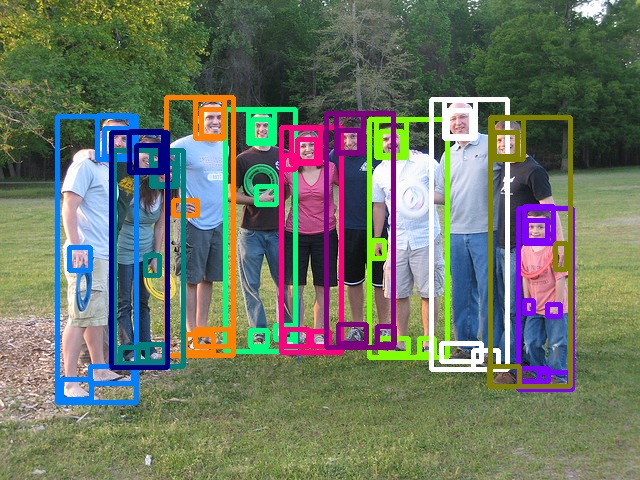}
	\caption{Qualitative results comparison of our BPJDet-L ({\bf right column}) for the joint {\it body-parts} detection task with the method Hier R-CNN (X101-FPN) \cite{yang2020hier} ({\bf left column}) on COCOHumanParts val-set images. These detected examples are directly fetched from its paper. The yellow arrows indicate wrongly associated cases by Hier R-CNN.}
	\label{BPJDetParts}
	\vspace{-10pt}
\end{figure}

Besides, for the subordinate relationship of body and parts, it can be observed that our models have only a slight decline under all categories AP$^{sub}$ metrics and have achieved discernible leadership in multiple sub indicators, reflecting advantages of our method in the body-parts association task. We also report body-head mMR$^{-2}$ results. Similarly, as results in CrowdHuman dataset shown in Table \ref{CrowdHumanHead}, BPJDet-L may not necessarily achieve a better association result than BPJDet-M due to its more detected challenging body and head instances. The difference is that the mMR$^{-2}$ values here are much lower, confirming the fact that the dataset COCOHumanParts is simpler than CrowdHuman which has a relative high population density. Finally, in Fig. \ref{BPJDetParts}, we show some comparative examples where detection results of our BPJDet-L are better than Hier-X101. We can intuitively see that our method can easily avoid some obviously unreasonable association pairs, which are troublesome to methods like Hier R-CNN with separate designs of detection and association.

\setlength{\tabcolsep}{1.5pt} %{0.32pt}
\begin{table*}[!t]\scriptsize % \small or \scriptsize or \tiny
	\begin{center}
	\caption{The performance comparison of the joint {\it body-parts} detection task in the val-set of COCOHumanParts. Results of all methods except ours are fetched from tables in the paper Hier R-CNN \cite{yang2020hier}. The marker $\star$ means using keypoints to guide the body-parts association. The marker $\dagger$ and $\ddag$ means using deformable convolutional layers (DCN) \cite{dai2017deformable} in backbone and multi-scale training with longer learning schedule (2$\times$), respectively. We mark the first-best and second-best results with red and green colors respectively.}
	\label{HumanParts}
	\vspace{-5pt}
	\begin{tabular}{l|c|ccccc|ccccccc|ccccc|c}
	\Xhline{1.2pt}
	\multirow{2}{*}{Methods} & \multirow{2}{*}{\makecell{Joint\\Detect?}} & \multicolumn{5}{c|}{All categories APs$\uparrow$} & \multicolumn{7}{c|}{Per categories APs$\uparrow$} & \multicolumn{5}{c|}{All categories APs (subordination)$\uparrow$} & \multirow{2}{*}{\makecell{body-head\\mMR$^{-2}\downarrow$}} \\
	\cline{3-19}
	~ & ~ & AP$_{.5:.95}$ & AP$_{.5}$ & AP$_{.75}$ & AP$_M$ & AP$_L$ & person & head & face & r-hand & l-hand & r-foot & l-foot & AP$_{.5:.95}^{sub}$ & AP$_{.5}^{sub}$ & AP$_{.75}^{sub}$ & AP$_M^{sub}$ & AP$_L^{sub}$ & ~ \\
	\Xhline{1.2pt}
	Faster-C4-R50 \cite{ren2015faster} & \xmark & 32.0 & 55.5 & 32.3 & 54.9 & 52.4 & 50.5 & 47.5 & 35.5 & 27.2 & 24.9 & 19.2 & 19.3 & --- & --- & --- & --- & --- & --- \\
	Faster-FPN-R50 \cite{he2017mask} & \cmark & 34.8 & 60.0 & 35.4 & 55.4 & 52.2 & 51.4 & 48.7 & 36.7 & 31.7 & 29.7 & 22.4 & 22.9 & 14.5 & 31.4 & 11.2 & 18.2 & 24.1 & ---\\
	Faster-FPN-R50$\star$ \cite{he2017mask} & \cmark & 34.8 & 60.0 & 35.4 & 55.4 & 52.2 & 51.4 & 48.7 & 36.7 & 31.7 & 29.7 & 22.4 & 22.9 & 19.1 & 38.5 & 16.0 & 22.4 & 33.6 & ---\\
	RetinaNet-R50 \cite{lin2017focal} & \xmark & 32.2 & 54.7 & 33.3 & 54.5 & 53.8 & 49.7 & 47.1 & 33.7 & 28.7 & 26.7 & 19.7 & 20.2 & --- & --- & --- & --- & --- & ---\\
	FCOS-R50 \cite{tian2019fcos} & \xmark & 34.1 & 58.6 & 34.5 & 55.1 & 55.1 & 51.1 & 45.7 & 40.0 & 29.8 & 28.1 & 22.2 & 21.9 & --- & --- & --- & --- & --- & ---\\
	\hline
	Faster-FPN-R101 \cite{he2017mask} & \xmark & 36.0 & 62.1 & 36.5 & 57.2 & 54.8 & 52.6 & 49.3 & 36.9 & 33.3 & 30.3 & 24.3 & 24.4 & --- & --- & --- & --- & --- & ---\\
	Faster-FPN-X101 \cite{he2017mask} & \xmark& 36.7 & 62.8 & 37.4 & 57.4 & 55.3 & 53.6 & 49.7 & 37.3 & 33.8 & 32.2 & 25.0 & 25.1 & --- & --- & --- & --- & --- & ---\\
	\hline
	Hier-R50 \cite{yang2020hier} & \cmark & 36.8 & 65.7 & 36.2 & 53.9 & 47.5 & 53.2 & 50.9 & 41.5 & 31.3 & 29.3 & 25.5 & 26.1 & 33.3 & 67.1 & 28.3 & 29.9 & 47.1 & ---\\
	Hier-R101 \cite{yang2020hier} & \xmark & 37.2 & 65.9 & 36.7 & 55.1 & 50.3 & 54.0 & 50.4 & 41.6 & 31.6 & 30.1 & 26.0 & 26.6 & --- & --- & --- & --- & ---  & ---\\
	Hier-X101 \cite{yang2020hier} & \cmark & 38.8 & 68.1 & 38.5 & 56.6 & 52.3 & 55.4 & 52.3 & 43.2 & 33.5 & 32.0 & 27.4 & 27.9 & 36.6 & 69.7 & 32.5 & 32.6 & 51.1 & ---\\
	\hline
	Hier-R50$\dagger$ \cite{yang2020hier} & \cmark & 38.6 & 67.9 & 38.3 & 55.9 & 51.7 & --- & --- & --- & --- & --- & --- & --- & 36.0 & 70.8 & 31.6 & 32.0 & 49.6 & ---\\
	Hier-R50$\ddag$ \cite{yang2020hier} & \cmark & 39.3 & 68.8 & 39.0 & 56.5 & 49.4 & --- & --- & --- & --- & --- & --- & --- & 37.3 & 70.8 & 33.5 & 33.1 & 52.0 & ---\\
	Hier-R50$\dagger\ddag$ \cite{yang2020hier} & \cmark& 40.6 & 70.1 & 40.7 & 57.5 & 51.5 & --- & --- & --- & --- & --- & --- & --- & 37.3 & \textcolor{red}{72.5} & 33.0 & 35.4 & 48.9 & ---\\
	\hline
	Hier-X101$\dagger$ \cite{yang2020hier} & \cmark & 40.3 & 70.1 & 40.1 & 58.2 & 53.6 & --- & --- & --- & --- & --- & --- & --- & 37.1 & 72.1 & 32.9 & 34.6 & 49.7 & ---\\
	Hier-X101$\ddag$ \cite{yang2020hier} & \cmark & 40.5 & 70.2 & 40.4 & 57.6 & 52.7 & --- & --- & --- & --- & --- & --- & --- & 38.6 & \textcolor{green}{72.3} & 35.0 & 35.2 & 53.5 & ---\\
	Hier-X101$\dagger\ddag$ \cite{yang2020hier} & \cmark & 42.0 & 71.6 & 42.3 & 59.0 & 53.3 & --- & --- & --- & --- & --- & --- & --- & 38.8 & \textcolor{green}{72.3} & 35.5 & 37.4 & 50.3 & --- \\
	\hline
	%BPJDet-S (Ours) & $\surd$ & 39.0 65.8 39.5 59.2 49.6 56.2 53.5 41.9 34.9 33.8 25.8 26.5 38.5 64.7 39.1 57.8 47.8 29.8 \\  % best.pt
	%BPJDet-M (Ours) & $\surd$ & 42.3 & 69.3 & 43.4 & 62.6 & 56.5 & 60.0 & 55.7 & 44.9 & 39.0 & 38.0 & 29.2 & 29.5 & 42.0 & 68.7 & 43.2 & 62.0 & 55.0 & 26.8 \\  % best.pt
	%BPJDet-L (Ours) & $\surd$ & 43.5 & 70.5 & 44.8 & 63.6 & 61.2 & 61.4 & 56.6 & 46.4 & 40.2 & 39.3 & 30.0 & 30.8 & 43.2 & 69.7 & 44.6 & 63.0 & 58.4 & 26.8 \\  % best.pt
	BPJDet-S (Ours) & \cmark & 38.9 & 65.5 & 39.4 & 59.1 & 49.7 & 56.3 & 53.5 & 41.9 & 34.7 & 33.7 & 25.5 & 26.5 & 38.4 & 64.4 & 38.9 & 57.7 & 47.5 & 29.8 \\  % best_mMR.pt
	BPJDet-M (Ours) & \cmark & 42.0 & 68.9 & 43.2 & 62.3 & 54.6 & 59.8 & 55.7 & 44.7 & 38.7 & 37.6 & 28.6 & 29.2 & 41.7 & 68.1 & 42.9 & 61.6 & 52.9 & \textcolor{green}{26.6} \\  % best_mMR.pt
	BPJDet-L (Ours) & \cmark & \textcolor{green}{43.6} & 70.6 & \textcolor{red}{45.1} & 63.8 & 61.8 & 61.3 & 56.6 & 46.3 & 40.4 & 39.6 & 30.2 & 30.9 & \textcolor{green}{43.3} & 69.8 & \textcolor{red}{44.7} & \textcolor{green}{63.2} & \textcolor{green}{58.9} & 26.8 \\  % best_mMR.pt
	BPJDet-L5u (Ours) & \cmark & \textcolor{red}{44.0} & \textcolor{red}{71.9} & \textcolor{green}{44.8} & \textcolor{red}{65.7} & \textcolor{red}{66.5} & 63.7 & 55.4 & 47.0 & 40.8 & 39.3 & 30.5 & 31.5 & \textcolor{red}{43.5} & 70.7 &\textcolor{green}{44.5} & \textcolor{red}{65.2} & \textcolor{red}{65.7} & \textcolor{red}{25.2} \\
	BPJDet-L8u (Ours) & \cmark & 43.5 & \textcolor{green}{71.7} & 44.2 & \textcolor{green}{65.4} & \textcolor{green}{63.1} & 60.6 & 55.3 & 47.1 & 40.5 & 39.4 & 30.4 & 31.2 & 41.6 & 68.6 & 42.4 & 59.9 & 55.1 & 29.7 \\
	\Xhline{1.2pt}
	\end{tabular}
	\end{center}
	\vspace{-10pt}
\end{table*}

%%%%%%%%%%%
\subsubsection{Body-Parts Joint Detection of Animals}\label{subsecBHJDA}
\begin{figure*}[]
	\includegraphics[height=0.339\columnwidth]{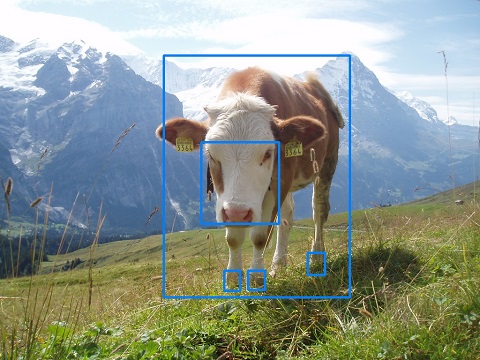}
	\includegraphics[height=0.339\columnwidth]{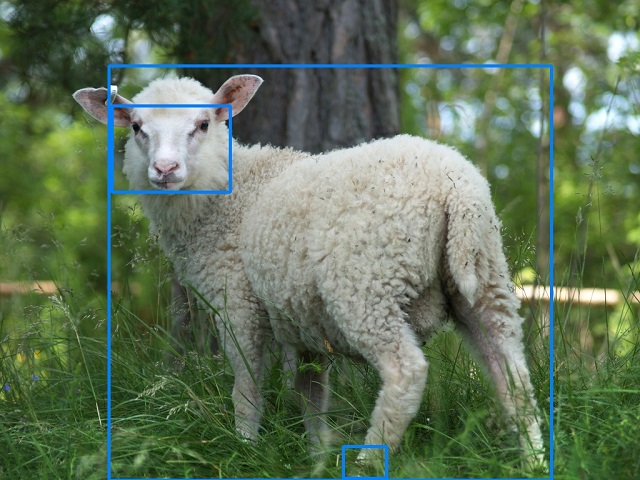}
	\includegraphics[height=0.339\columnwidth]{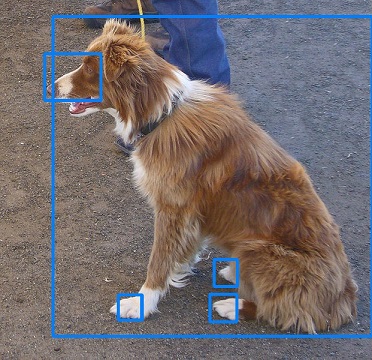}
	\includegraphics[height=0.339\columnwidth]{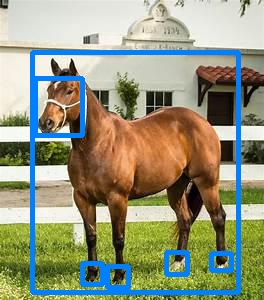}
	\includegraphics[height=0.339\columnwidth]{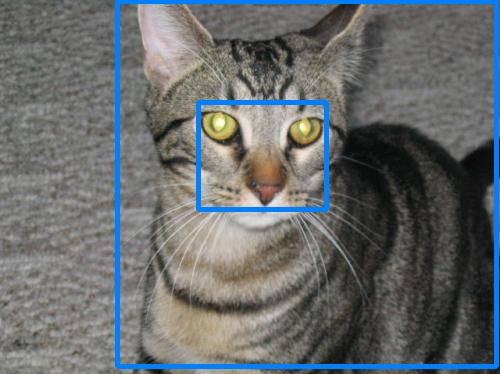}\\
	\vspace{-9pt}\\
	\includegraphics[height=0.275\columnwidth]{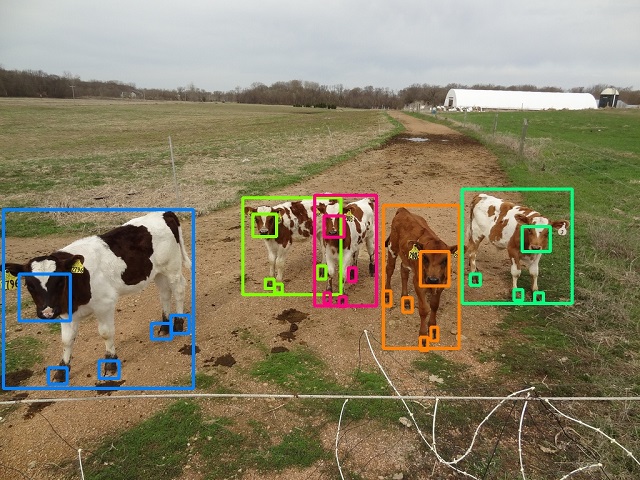}
	\includegraphics[height=0.275\columnwidth]{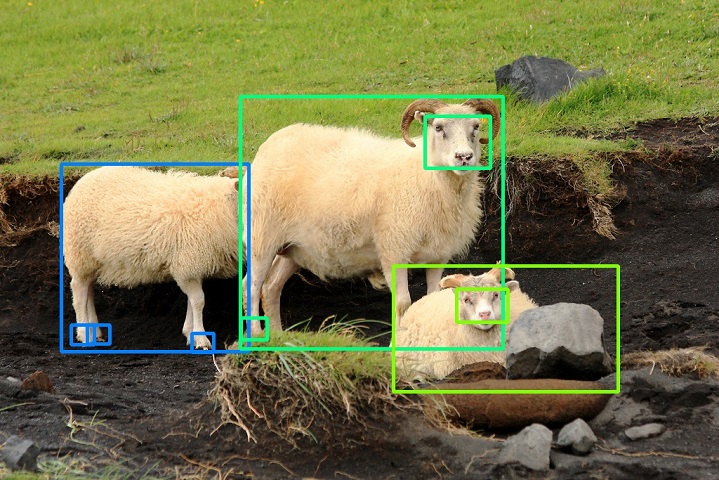}
	\includegraphics[height=0.275\columnwidth]{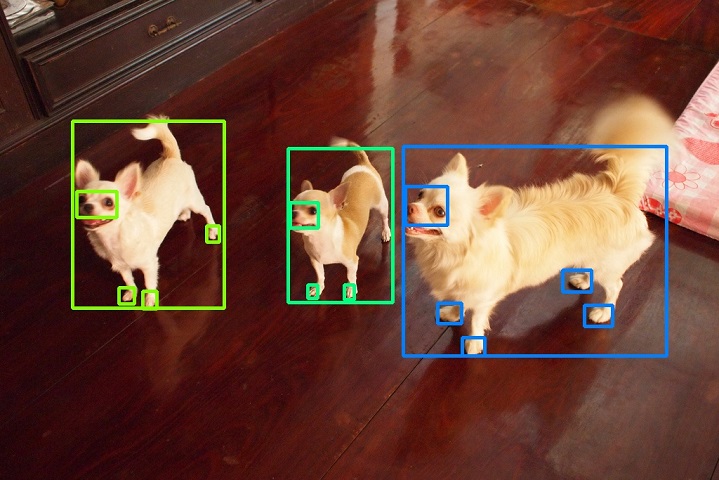}
	\includegraphics[height=0.275\columnwidth]{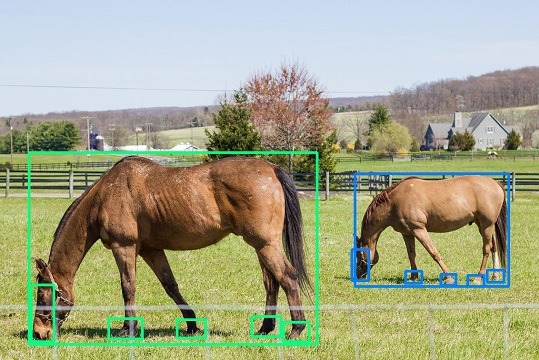}
	\includegraphics[height=0.275\columnwidth]{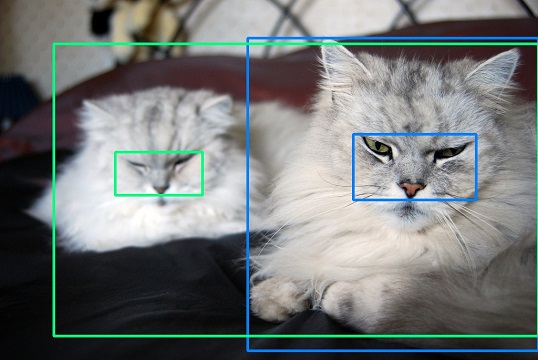}
	\caption{Qualitative results of our BPJDet-L for the joint {\it body-parts} detection task of quadruped animals (e.g., dog, cat, sheep, horse and cow).}
	\label{BPJDetAnimals}
	\vspace{-10pt}
\end{figure*}

For exploratory purposes, we choose the reconstructed dataset Animals5C to conduct the body-parts joint detection of animals. Due to the fact that the GT boxes of each part of an animal are automatically generated through 2D keypoints and body size, they do not tightly surround the object. Therefore, although each loss in BPJDet can converge in training, the final APs are not satisfactory. Nevertheless, we can quantitatively confirm that BPJDet does work with producing relative high results of precision and recall (about {\bf 80}\% in the val-set). In addition, as shown in Fig. \ref{BPJDetAnimals}, our qualitative results on images from the val-set also demonstrate that the body-parts joint detection of animals is successful. These results indicate that our tentative generalization testing of BPJDet on quadruped animals is persuasive and meaningful.

%%%%%%%%%%%%%%%%%%%%%%%%
\subsection{Benefits of Generic BPJDet}

Associating each detected part (e.g., head and hand) to a human body is beneficial for many downstream tasks. This subsection demonstrates the benefits of this ability by our proposed generic BPJDet for two such tasks: accurate crowd head detection and hand contact estimation. Without losing generality, we only train BPJDet using anchor-based detectors in these applications.

\subsubsection{Body-Head for Accurate Crowd Head Detection}\label{subsecBHfACHD}
\begin{figure}[]
	\includegraphics[width=0.495\columnwidth]{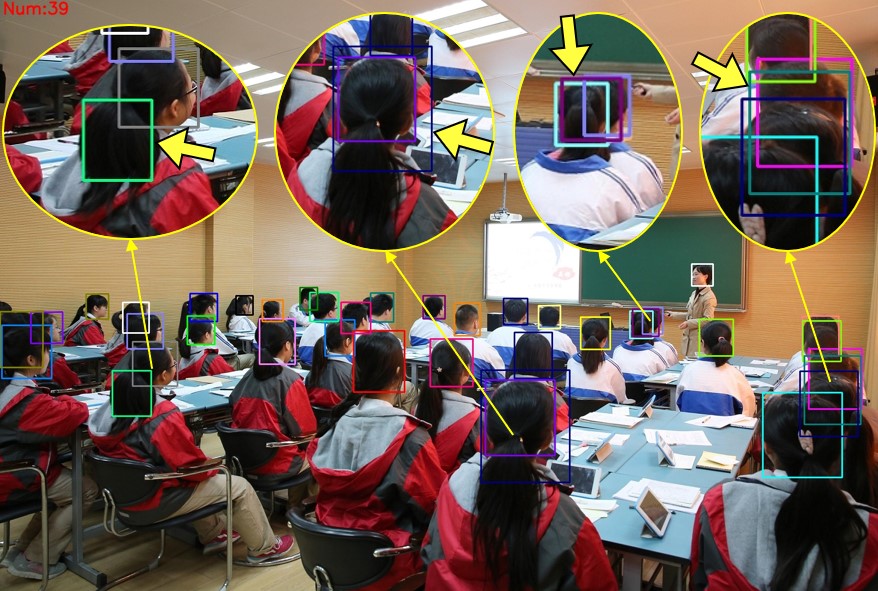}
	\includegraphics[width=0.495\columnwidth]{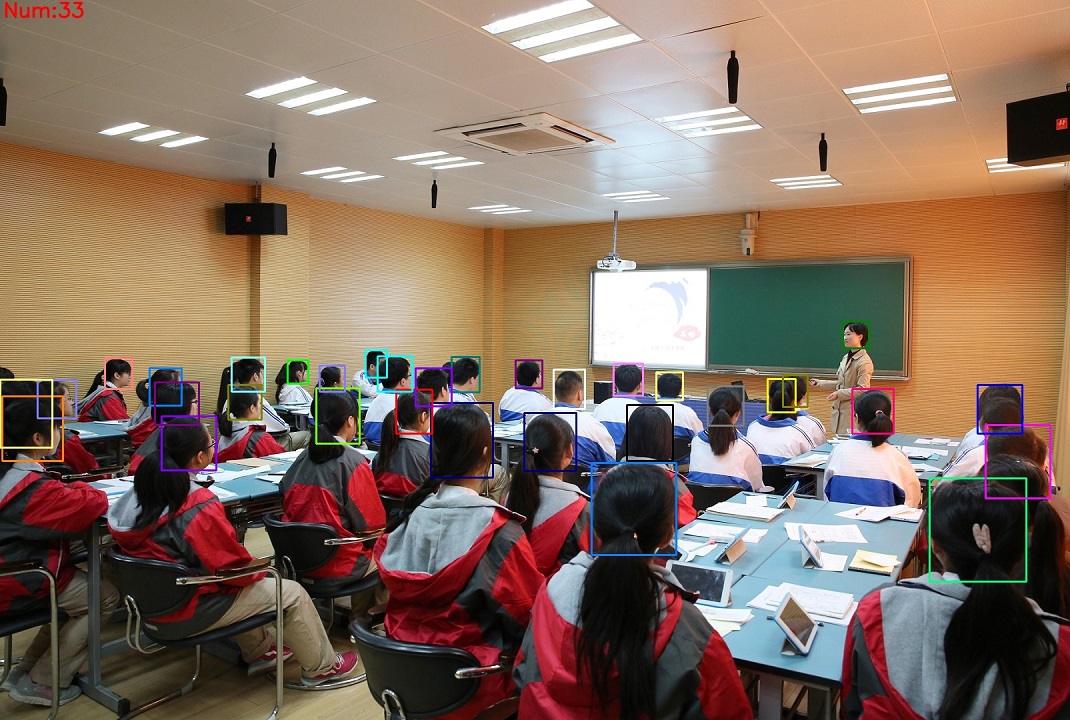}\\
	\vspace{-9pt}\\
	\includegraphics[width=0.495\columnwidth]{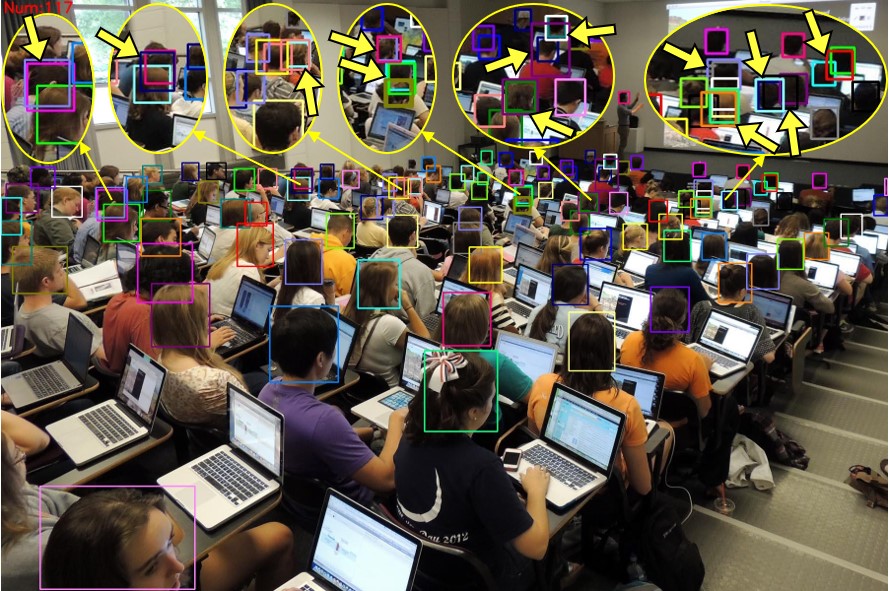}
	\includegraphics[width=0.495\columnwidth]{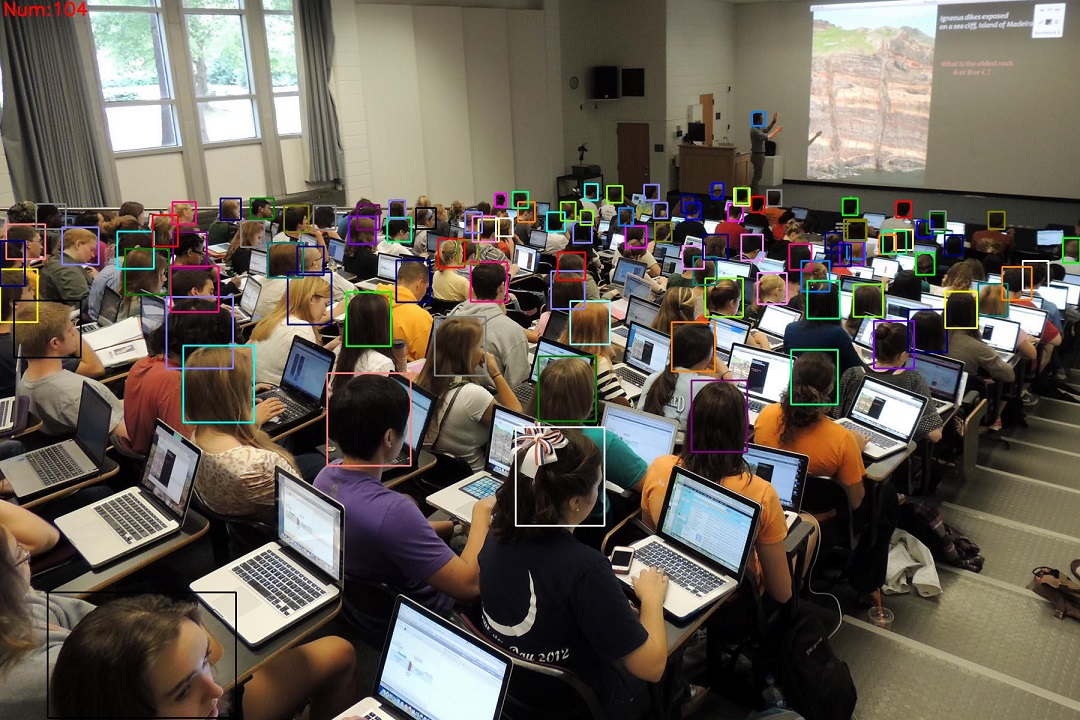}\\
	\vspace{-9pt}\\
	\includegraphics[width=0.495\columnwidth]{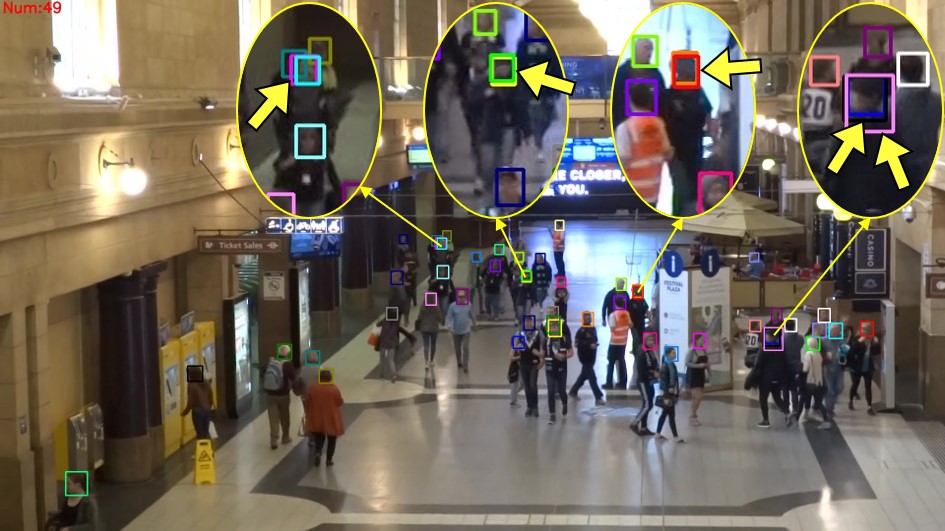}
 	\includegraphics[width=0.495\columnwidth]{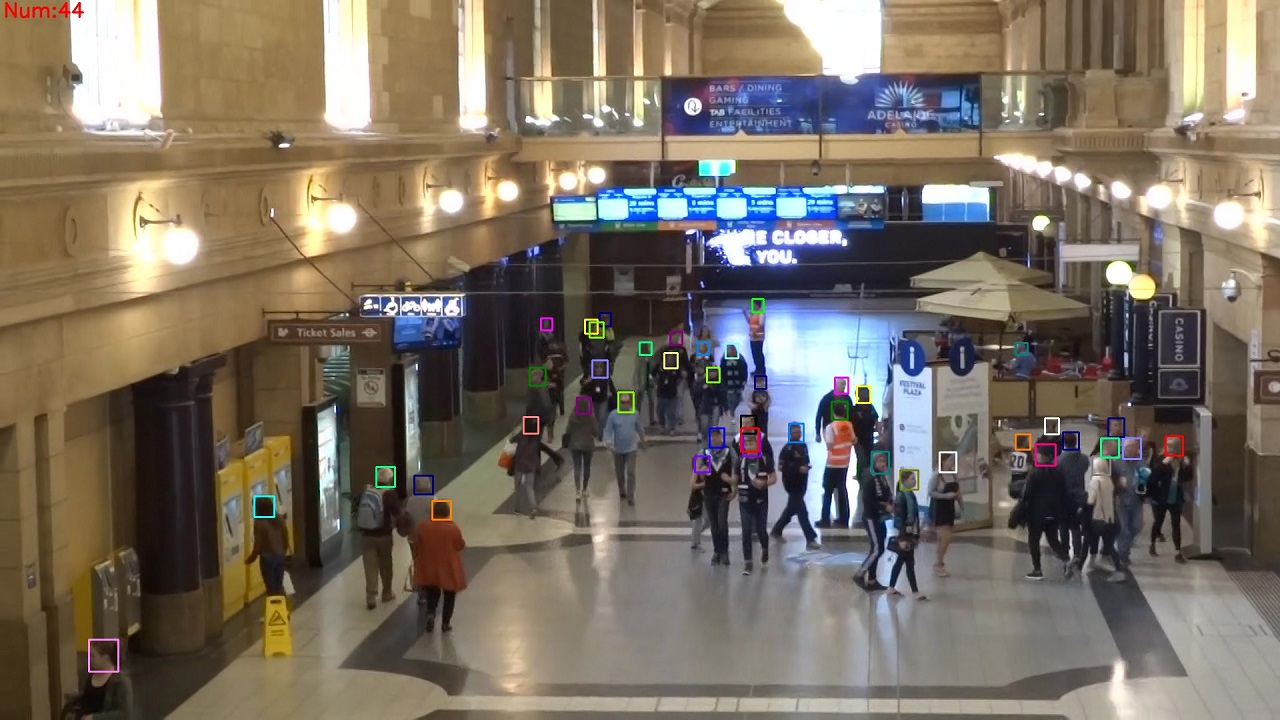}
	\vspace{-9pt}\\
	\includegraphics[width=0.495\columnwidth]{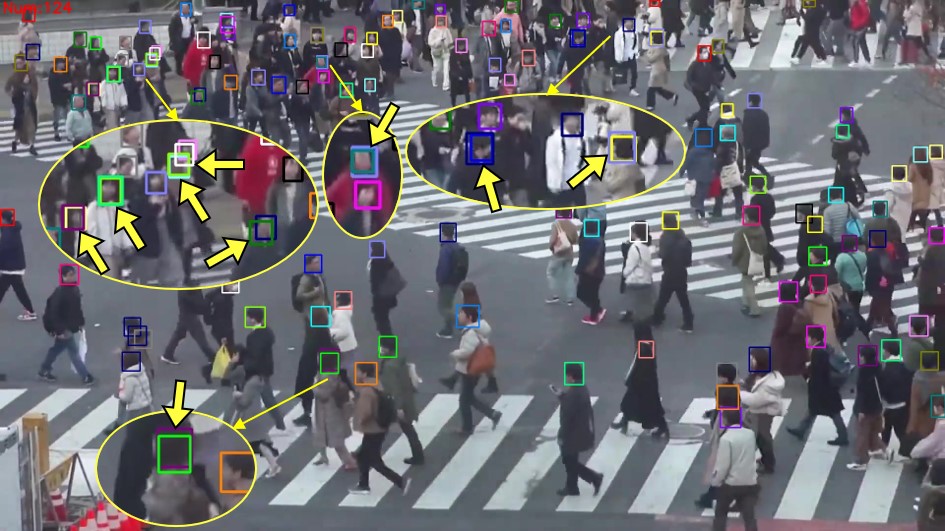}
 	\includegraphics[width=0.495\columnwidth]{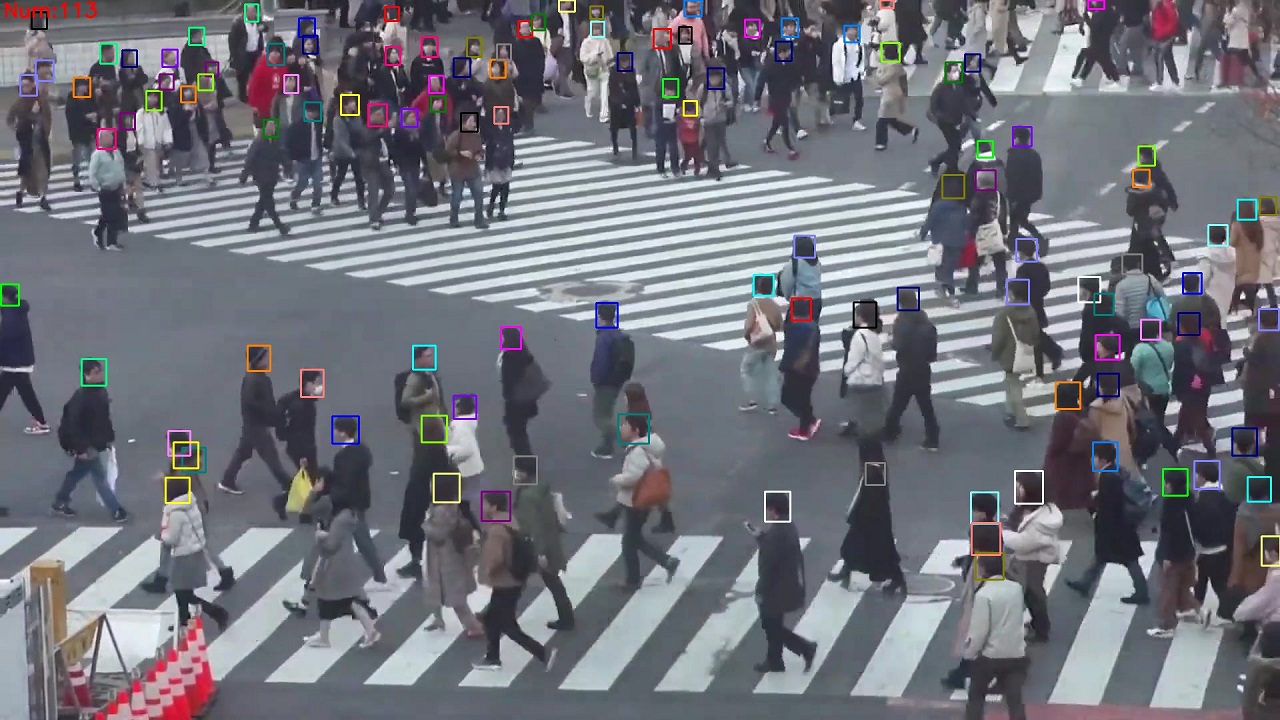}
	\caption{Qualitative results comparison of BPJDet-L trained on CrowdHuman, and directly tested on unseen target domains. The top and bottom two images are from SCUT Head dataset \cite{peng2018detecting} and CroHD train-set \cite{sundararaman2021tracking}, respectively. The {\bf left column} and {\bf right column} is predicted by BPJDet-L without using and with using body-head associations, respectively. The yellow arrows indicate false positive head boxes, which are greatly reduced by using body-head associations. Please zoom in corresponding areas to check in detail.}
	\label{appCrowdVis}
	\vspace{-10pt}
\end{figure}

As discussed in Section \ref{datasetsIntro}, the chosen two datasets SCUT Head dataset \cite{peng2018detecting} and CroHD train-set \cite{sundararaman2021tracking} are not suitable for retraining or finetuning BPJDet. Although the larger in-the-wild CrowdHuman dataset has similar congested populations comparing to them, they mutually have scene-level inter domain differences. Thus, we attempt to conduct cross domain generalization experiments on these two target datasets with specific domains. All results are summarized in Table \ref{appCrowdTab} and Fig. \ref{appCrowdVis}. We also report non-cross-domain evaluation results on the CrowdHuman val-set.

\setlength{\tabcolsep}{4.2pt} %{0.32pt}
\begin{table}[!t] % \small or \scriptsize or \tiny
	\begin{center}
	\caption{The performance comparison experiments of accurate crowd head detection. The $\ddag$ means setting a NMS confidence threshold as 0.5. We mark the metric variations of BPJDet-S$\ddag$ for using a higher threshold 0.5 in the upper right. The $\dagger$ means not applying body-head associations to double check head boxes. We mark the metric variations of BPJDet-S/M/L after using body-head associations in the lower right.}
	\label{appCrowdTab}
	\vspace{-5pt}
	\begin{tabular}{l|l|llll}
	\Xhline{1.2pt}
	\multirow{2}{*}{Datasets} & \multirow{2}{*}{Methods} & \multicolumn{4}{c}{Detection Metrics$\uparrow$} \\
	\cline{3-6}
	~ & ~ & Precision & Recall & F1 Score & mAP \\
	\Xhline{1.2pt}
	\multirow{7}{*}{\makecell{Crowd-\\Human\\ val-set\\ \cite{shao2018crowdhuman}}} & BPJDet-S$\ddag$ & 90.6$^{+20.9}$ & 74.2$^{-9.3}$ & 81.6$^{-0.4}$ & 72.5$^{-7.9}$ \\  % not using association, conf-thre=0.5
	\cline{2-6}
	~ & BPJDet-S$\dagger$ & 69.7 & 83.5 & 82.0 & 80.2 \\  % not using association
	~ & BPJDet-M$\dagger$ & 72.1 & 83.7 & 82.5 & 80.7 \\  % not using association
	~ & BPJDet-L$\dagger$ & 76.6 & 81.6 & 82.0 & 78.6 \\  % not using association
	\cline{2-6}
	~ & BPJDet-S & 81.1$_{+10.4}$ & 81.7$_{-1.8}$ & 82.6$_{+0.6}$ & 79.0$_{-1.2}$ \\  % using association
	~ & BPJDet-M & 82.6$_{+10.5}$ & 81.9$_{-1.8}$ & 83.1$_{+0.6}$ & 79.5$_{-1.2}$ \\  % using association
	~ & BPJDet-L & 85.5$_{+8.9}$ & 80.3$_{-1.3}$ & 83.1$_{+1.1}$ & 77.9$_{-0.7}$ \\  % using association
	\Xhline{0.8pt}
	\multirow{7}{*}{\makecell{SCUT\\Head\\Part\_B\\ \cite{peng2018detecting}}} & BPJDet-S$\ddag$ & 94.0$^{+9.5}$ & 76.7$^{-6.7}$ & 84.5$^{-0.8}$ & 75.8$^{-6.0}$ \\  % not using association, conf-thre=0.5
	\cline{2-6}
	~ & BPJDet-S$\dagger$ & 84.5 & 83.4 & 85.3 & 81.8 \\  % not using association
	~ & BPJDet-M$\dagger$ & 84.2 & 82.8 & 84.8 & 81.2 \\  % not using association
	~ & BPJDet-L$\dagger$ & 85.1 & 80.4 & 84.3 & 79.1 \\  % not using association
	\cline{2-6}
	~ & BPJDet-S & 90.4$_{+5.9}$ & 81.7$_{-1.7}$ & 85.9$_{+0.6}$ & 80.5$_{-1.3}$ \\  % using association
	~ & BPJDet-M & 90.6$_{+6.4}$ & 80.9$_{-1.9}$ & 85.5$_{+0.7}$ & 79.6$_{-1.6}$ \\  % using association
	~ & BPJDet-L & 91.7$_{+6.6}$ & 79.3$_{-1.1}$ & 85.1$_{+0.8}$ & 78.2$_{-0.9}$ \\  % using association
	\Xhline{0.8pt}
	\multirow{7}{*}{\makecell{CroHD\\train-set\\ \cite{sundararaman2021tracking}}} & BPJDet-S$\ddag$ & 67.8$^{+17.9}$ & 35.2$^{-14.3}$ & 46.3$^{-4.8}$ & 27.1$^{-8.8}$ \\  % not using association, conf-thre=0.5
	\cline{2-6}
	~ & BPJDet-S$\dagger$ & 49.9 & 49.5 & 51.1 & 35.9 \\  % not using association
	~ & BPJDet-M$\dagger$ & 53.3 & 46.3 & 50.0 & 33.1 \\  % not using association
	~ & BPJDet-L$\dagger$ & 58.2 & 42.2 & 49.0 & 31.3 \\  % not using association
	\cline{2-6}
	~ & BPJDet-S & 61.5$_{+11.6}$ & 46.8$_{-2.7}$ & 53.2$_{+2.1}$ & 34.7$_{-1.2}$ \\  % using association
	~ & BPJDet-M & 62.4$_{+9.2}$ & 44.2$_{-2.1}$ & 51.7$_{+1.7}$ & 32.1$_{-1.0}$ \\  % using association
	~ & BPJDet-L & 64.3$_{+6.1}$ & 41.2$_{-1.0}$ & 50.2$_{+1.2}$ & 30.9$_{-0.4}$ \\  % using association
	\Xhline{1.2pt}
	\end{tabular}
	\end{center}
	\vspace{-15pt}
\end{table}

Specifically, to evaluate BPJDet trained on Crowdhuman directly on two unknown target domains, we need to lower the NMS confidence threshold to better generalize, rather than using a larger threshold (e.g., 0.5 or 0.9) as in supervised learning tests. Without losing generality, we here set confidence thresholds as 0.2 for NMS filtering. However, simply relying on lower thresholds is not robust, which will lead to more candidates being obtained, while also introducing more false positive predictions. This means better recall rates, F1 scores and mAP, but worse precision. We can check related metric variations of BPJDet-S$\ddag$ ($\tau^b_{conf}\!=\!\tau^p_{conf}\!=\!0.5$) and BPJDet-S$\dagger$ ($\tau^b_{conf}\!=\!\tau^p_{conf}\!=\!0.2$) on three datasets in Table \ref{appCrowdTab} for more details. Despite improvements in recall, F1, and mAP values, the precision is rapidly declining. More qualitative visualization of false positives for cross domain testing is shown in the left column of Fig. \ref{appCrowdVis}.

Fortunately, our proposed BPJDet can predict body-head pairs rather than just apply head detection. To reduce false positives, we can utilize these pairs to perform a double check, retaining only candidates who detect body and head. Using this strategy, we can greatly improve precision and slightly increase F1 scores with minimal recall and mAP decreases in three test sets, as shown in Table \ref{appCrowdTab} (BPJDet-S$\dagger$/M$\dagger$/L$\dagger$ vs. BPJDet-S/M/L). Qualitative comparison is shown in Fig. \ref{appCrowdVis}. All these impressive improvements benefit from the reliable body-head association ability of BPJDet. This experiment has explored our proposed BPJDet with {\it body-head} joint detection models for realizing stable, robust and accurate human head detection in crowded scenes. It also demonstrates the great potential of BPJDet for cross domain generalization in head detection, which is meaningful to following in-the-wild applications such as surveillance, crowd counting and general head tracking.

%%%%%%%%%%%
%ContactHands: Detecting hands and recognizing physical contact in the wild \cite{narasimhaswamy2020detecting}
% augmented reality and virtual reality
\subsubsection{Body-Hand for Hand Contact Estimation}\label{subsecBHfHCE}
Following the test setting of physical contact analysis in BodyHands \cite{narasimhaswamy2022whose}, we here explore the benefits of body-hand joint detection for recognizing or estimating the physical contact state of a hand on dataset ContactHands \cite{narasimhaswamy2020detecting}. Hand contact estimation is a high-level task that has many further applications such as human understanding, activity recognition and AR/VR. Generic hand contact estimation is a complex problem. ContactHands has defined four contact states: No-Contact, Self-Contact, Person-Contact and Object-Contact. They are not mutually exclusive considering multiple states of one hand. And the Person-Contact category is the most challenging one due to the difficulty of distinguishing it with Self-Contact. It is nontrivial to determine if one hand is part of the same body (Self-Contact) or a different body (Person-Contact). By utilizing the visual appearance of a hand and its surrounding local context, BodyHands \cite{narasimhaswamy2022whose} proposed two approaches as listed in Table \ref{appContactTab} to exclusively improve the baseline of Person-Contact estimation by leveraging the body-hand association. Methods in \cite{narasimhaswamy2020detecting, narasimhaswamy2022whose} are all based on ResNet-101.

\setlength{\tabcolsep}{2.7pt} %{0.4pt}
\begin{table}[!t]\small  % \small or \scriptsize or \tiny
	\begin{center}
	\caption{The performance comparison of hand contact estimation. The states NC, SC, PC and OC denotes No-Contact, Self-Contact, Person-Contact and Object-Contact, respectively. The marker $\star$ means using the extra video-frame training dataset 100DOH \cite{shan2020understanding}. We can further advance the SOTA thanks to the superior body-hand association ability of BPJDet.}
	\label{appContactTab}
	\vspace{-5pt}
	\begin{tabular}{l|ccccc}
	\Xhline{1.2pt}
	Methods & AP$_{NC}$ & AP$_{SC}$ & AP$_{PC}$ & AP$_{OC}$ & mAP \\
	\Xhline{1.2pt}
	Mask R-CNN \cite{he2017mask, narasimhaswamy2020detecting} & 60.52 & 51.62 & 33.79 & 67.43 & 53.31 \\
	ContactHands \cite{narasimhaswamy2020detecting} & 62.48 & 54.31 & 39.51 & 73.34 & 57.41 \\
	Heuristic Method \cite{narasimhaswamy2022whose} & 62.48 & 54.31 & 40.89 & 73.34 & 57.56 \\
	ContactHands$\star$ \cite{narasimhaswamy2020detecting} & 63.90 & 59.30 & 42.01 & 70.49 & 58.93 \\
	End-to-end Method \cite{narasimhaswamy2022whose} & 64.74 & 56.12 & 47.09 & 74.32 & 60.56 \\
	\hline
	BPJDetPlus-S (Ours) & 58.87 & 54.69 & 45.51 & 66.62 & 56.42 \\
	BPJDetPlus-M (Ours) & 62.67 & 56.16 & 46.33 & 67.48 & 58.16 \\
	BPJDetPlus-L (Ours) & 62.98 & 57.39 & 50.32 & 69.55 & 60.06 \\
	\Xhline{1.2pt}
	\end{tabular}
	\end{center}
	\vspace{-10pt}
\end{table}

\setlength{\tabcolsep}{2.1pt} %{0.4pt}
\begin{table}[!t]\scriptsize  % \small or \scriptsize or \tiny
	\begin{center}
	\caption{The influence of contact state weights $w_h$ and $w_b$ for the hand and body instances on various AP values of the trained BPJDetPlus-S model.}
	\label{handbodyTradeOff}
	\vspace{-5pt}
	\begin{tabular}{l|ccccccccccc}
	\Xhline{1.2pt}
	$w_h$ & 0.0 & 0.1 & 0.2 & 0.3 & 0.4 & 0.5 & 0.6 & 0.7 & 0.8 & 0.9 & 1.0 \\
	\hline
	$w_b$ & 1.0 & 0.9 & 0.8 & 0.7 & 0.6 & 0.5 & 0.4 & 0.3 & 0.2 & 0.1 & 0.0 \\
	\Xhline{1.2pt}
	AP$_{NC}$ & 55.51 & 56.39 & 56.99 & 57.58 & 58.02 & 58.39 & 58.87 & 58.75 & 58.85 & {\bf 58.89} & 58.76 \\
	AP$_{SC}$ & 52.18 & 52.85 & 53.38 & 53.91 & 54.32 & 54.51 & 54.69 & 54.85 & 54.83 & 54.76 & {\bf 54.91} \\
	AP$_{PC}$ & 38.60 & 41.14 & 42.66 & 43.29 & 43.87 & 44.90 & {\bf 45.51} & 45.00 & 44.27 & 44.22 & 43.39 \\
	AP$_{OC}$ & 63.30 & 65.02 & 65.60 & 65.94 & 66.23 & 66.48 & 66.62 & {\bf 66.66} & 66.65 & 66.50 & 66.38 \\
	mAP & 52.40 & 53.85 & 54.66 & 55.18 & 55.61 & 56.07 & {\bf 56.42} & 56.32 & 56.15 & 56.09 & 55.86 \\ 
	\Xhline{1.2pt}
	\end{tabular}
	\end{center}
	\vspace{-10pt}
\end{table}

\begin{figure}[]
	\includegraphics[width=0.495\columnwidth]{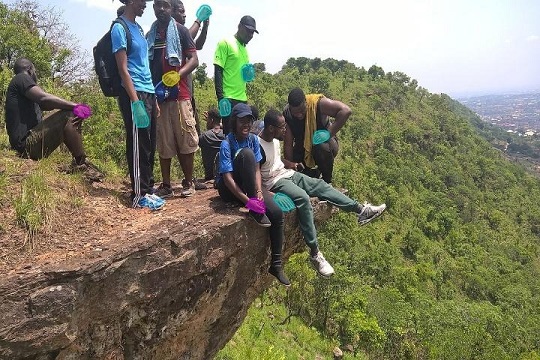}
	\includegraphics[width=0.495\columnwidth]{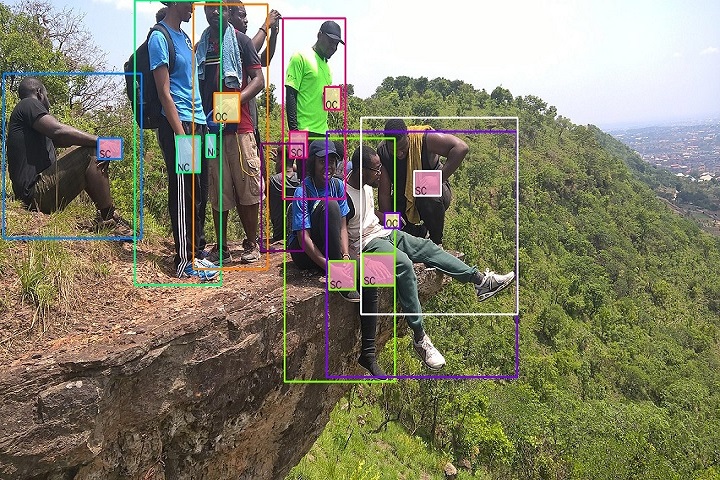}\\
	\vspace{-9pt}\\
	\includegraphics[width=0.495\columnwidth]{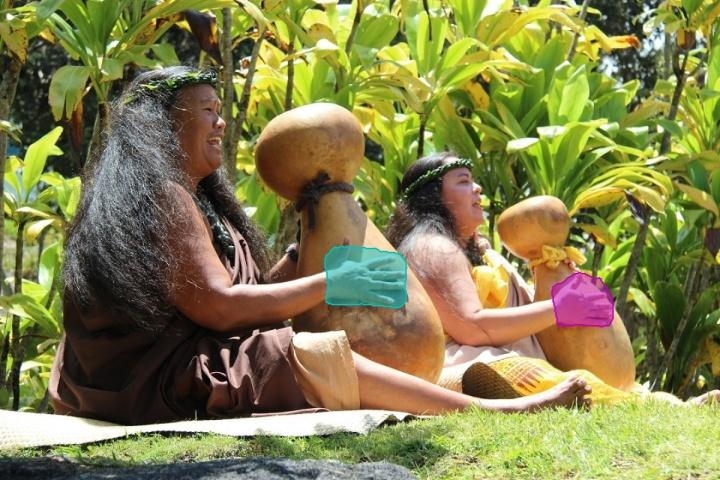}
	\includegraphics[width=0.495\columnwidth]{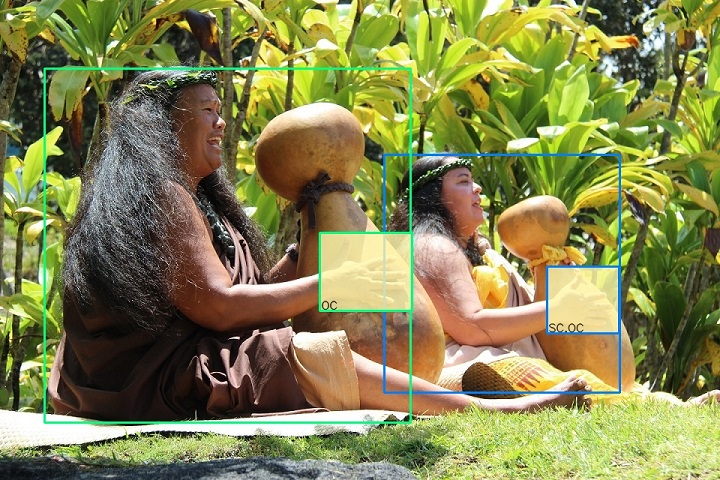}\\
	\vspace{-9pt}\\
	\includegraphics[width=0.495\columnwidth]{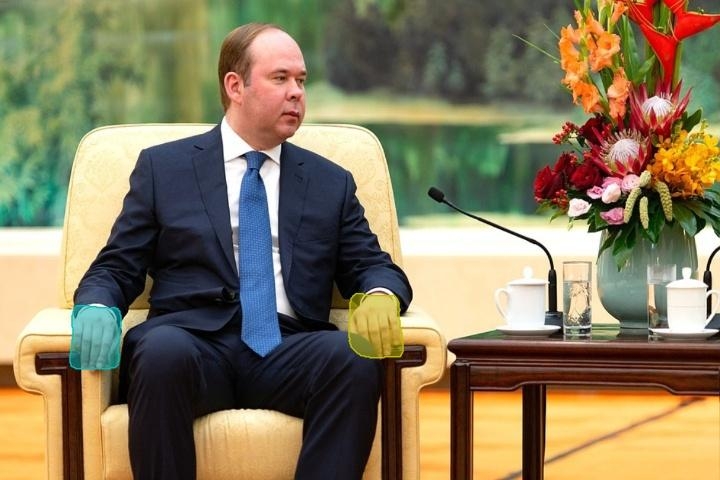}
	\includegraphics[width=0.495\columnwidth]{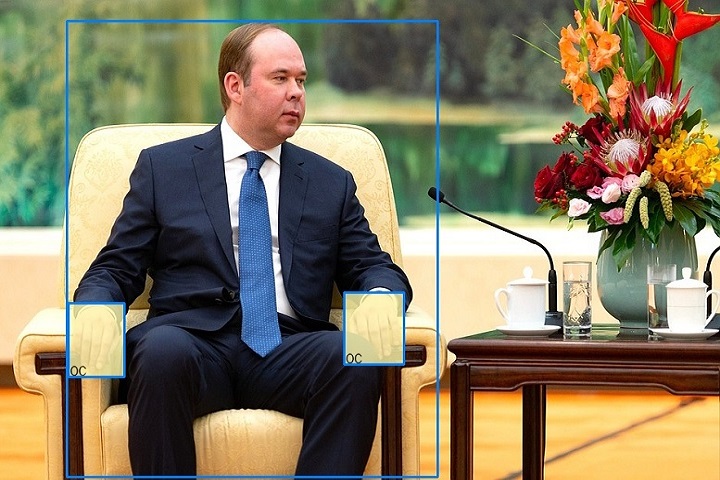}\\
	\vspace{-9pt}\\
	\includegraphics[width=0.495\columnwidth]{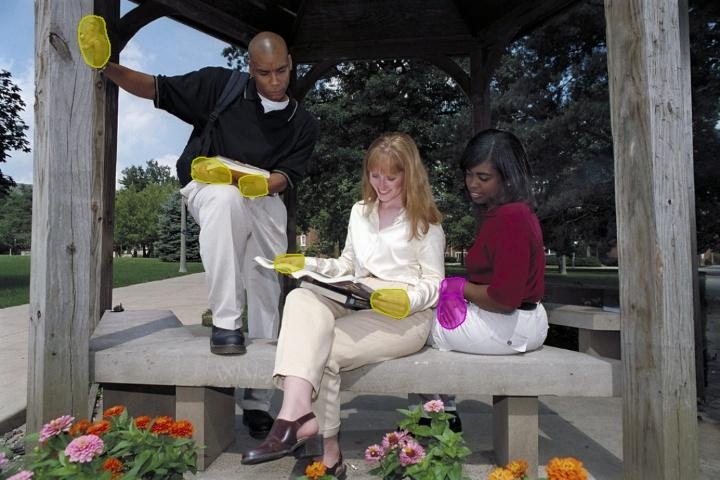}
	\includegraphics[width=0.495\columnwidth]{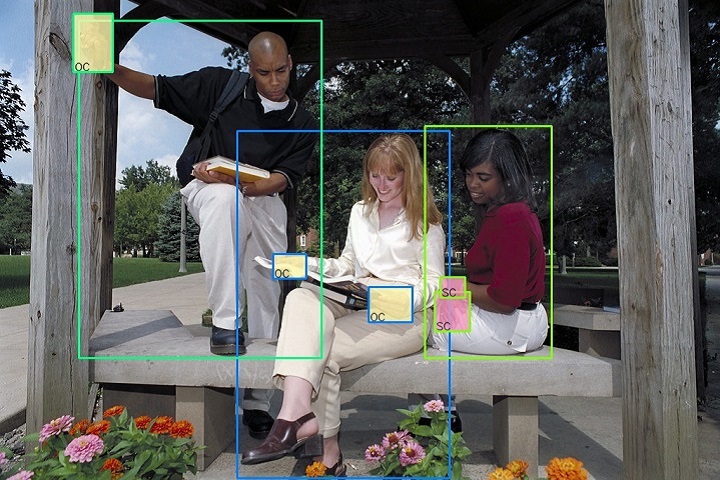}\\
	\vspace{-9pt}\\
	\includegraphics[width=1.0\columnwidth]{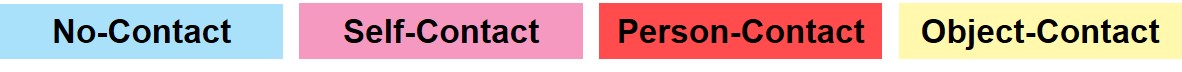}
	\caption{Qualitative results comparison of BPJDetPlus-L ({\bf right column}) and the SOTA method in ContactHands \cite{narasimhaswamy2020detecting} based on Mask R-CNN \cite{he2017mask} ({\bf left column}). All failure cases are fetched from the original paper \cite{narasimhaswamy2020detecting}. We visualize detected hand instances by their predicted contact state color. Please zoom in corresponding areas to check.}
	\label{appContactVis}
	\vspace{-10pt}
\end{figure}

Similarly, we implemented our BPJDetPlus on ContactHands as explained in Section \ref{BPJDetPlus}. As in Table \ref{handbodyTradeOff}, we empirically assign weights $w_h$ and $w_b$ of the hand and body instances as 0.6 and 0.4 by offline searching based on the trained BPJDetPlus-S model. Relying solely on tight hand instances ($w_h\!=\!1.0$) or loose body instances ($w_b\!=\!1.0$) to estimate the hand contact state is insufficient and not optimal. Finally, we summarize our results in Table \ref{appContactTab}. Without the need for meticulously designed attention mechanisms or many additional tuning steps like in \cite{narasimhaswamy2020detecting, narasimhaswamy2022whose}, our method BPJDetPlus-L improves the most challenging AP$_{PC}$ from the best 47.09\% to {\bf 50.32\%}, which profoundly benefits from the stronger {\it body-hand} association ability. The largest gap appears in AP$_{OC}$ values (69.55\% vs. 74.32\%) due to the applying of a pre-trained universal object detector to detect all other common objects in the image in methods \cite{narasimhaswamy2020detecting, narasimhaswamy2022whose}. Nonetheless, we achieve a comparable mAP {\bf 60.06\%} with the SOTA 60.56\%. More qualitative results comparison of our method with the SOTA method in ContactHands \cite{narasimhaswamy2020detecting} is shown in Fig. \ref{appContactVis}. These results certificate the strong adaptability of our proposed extended representation and effortless performance gains of using body-hand pairs for the hand contact estimation task.

%%%%%%%%%%%%%%%%%%%%%%%%%%%%%%%%%%%%%%%%%%%%%%%%%%%%%%%%%%%%%%%
\section{Discussion and Conclusion}

In this paper, we pointed out many problems with existing body-part joint detection methods, including inefficient separated branches of detection and association, reliance on heuristic rules to design association modules, and addiction to outdated detectors. Compared to some popular human-centered topics such as pedestrian detection, face detection and pose estimation, the joint detection of body-part is less-explored. We thus thoroughly investigated technical difficulties related to body-part joint detection and its potential to assist downstream tasks. To promote this community, we propose a novel generic single-stage body-part joint detector named BPJDet to address the challenging paired object detection and association problem. Inspired by center-offset regression in general object detection and human pose estimation, we extend the traditional object representation by appending body-part displacements, and design favorable multi-loss functions to enable joint training of detection and association tasks. Furthermore, BPJDet is not limited to a specific or single body part. And it is also not picky about using anchor-based or anchor-free detectors. Quantitative SOTA results and impressive qualitative performance on four public body-part or body-parts datasets of human as well as one body-parts dataset of quadruped animals suffice to demonstrate the robustness, adaptability and superiority of our proposed BPJDet. Besides, we have conducted tests on two closely related downstream applications, including accurate crowd head detection and hand contact estimation, to further verify obvious benefits of the advanced body-part association ability of BPJDet.

In summary, we envision that our BPJDet can serve as a strong baseline on body-part joint detection benchmarks and may benefit many other applications. Nonetheless, the performance of BPJDet in some difficult scenarios is not satisfactory, as shown of some missed and failure cases in Fig. \ref{BPJDetHand}, \ref{BPJDetParts} and \ref{appContactVis}. We assume that these challenges may be alleviated by utilizing more powerful backbones such as transformer-based DETRs, applying unsupervised learning with large-scale unlabeled images, or exploring advanced strategies for more robust multi-task joint learning. In addition, body-part pairs can help with many trendy tasks, such as stable multi-person/head tracking, fine-grained human-object interaction (HOI) and generic human-object contact detection. We will study them in the future.

%%%%%%%%%%%%%%%%%%%%%%%%%%%%%%%%%%%%%%%%%%%%%%%%%%%%%%%%%%%%%%%%%%%

\section*{Acknowledgments}
We acknowledge the effort from authors of human-related datasets including CityPersons, CrowdHuman, BodyHands, COCOHumanParts, SCUT Head Part\_B, CroHD and ContactHands. These datasets make researches and downstream applications about generic body-part joint detection and association possible. 

%%%%%%%%%%%%%%%%%%%%%%%%%%%%%%%%%%%%%%%%%%%%%%%%%%%%%%%%%%%%%%%

%\vfill\pagebreak


\bibliographystyle{IEEEtran}
\bibliography{refs}
%{\small \bibliography{refs}}
%{\footnotesize \bibliography{refs}}

\end{document}